# Snail Homing and Mating Search Algorithm: A Novel Bio-Inspired Metaheuristic Algorithm

Anand J Kulkarni[1], Ishaan R Kale[1], Apoorva Shastri[1], Aayush Khandekar[2]

[1]Institute of Artificial Intelligence, Dr Vishwanath Karad MIT World Peace University, 124 Paud Road, Kothrud, Pune, MH 411038, India

[2]Stevens School of Business, Stevens Institute of Technology, NJ 07030, USA

anand.j.kulkarni@mitwpu.edu.in; ishaan.kale@mitwpu.edu.in; apoorva.shastri@mitwpu.edu.in; akhandek1@stevens.edu

**Abstract**

In this paper, a novel Snail Homing and Mating Search (SHMS) algorithm is proposed. It is inspired from the biological behaviour of the snails. Snails continuously travels to find food and a mate, leaving behind a trail of mucus that serves as a guide for their return. Snails tend to navigate by following the available trails on the ground and responding to cues from nearby shelter homes. The proposed SHMS algorithm is investigated by solving several unimodal and multimodal functions. The solutions are validated using standard statistical tests such as two-sided and pairwise signed rank Wilcoxon test and Friedman rank test. The solution obtained from the SHMS algorithm exhibited superior robustness as well as search space exploration capabilities within the less computational cost. The real-world application of SHMS algorithm is successfully demonstrated in the engineering design domain by solving three cases of design and economic optimization shell and tube heat exchanger problem. The objective function value and other statistical results obtained using SHMS algorithm are compared with other well-known metaheuristic algorithms.

## 1. Introduction

Several nature inspired metaheuristic algorithms so far have been proposed by the researchers. These algorithms are classified into several sub-domains (Kale and Kulkarni, 2021) such as (i) biology/ evolutionary algorithms, (ii) swarm algorithms, (iii) socio inspired algorithms and (vi) physics and chemical based algorithms. Evolutionary Algorithms (EAs) mimics the evolution of several biological species. EAs are derived using the operators such as crossover and mutation. The most commonly used EA is Genetic Algorithm (GA) (Holland, 1992) used these operators to generate the optimum solution. There are several other algorithms like Differential Evolution, Evolutionary Programming (Fogel et al., 1966), Genetic Programming, Evolutionary Strategies (Michalewicz and Schoenauer, 1996). The concept of swarm optimization algorithm was adopted from the literature proposed by Happener and Grenarder (1990). Based on this, Kennedy and Eberhert (1995) proposed Particle Swarm

Optimization (PSO). It mimics the foraging behavior of flock of birds/ school of fishes. The other well-known swarm algorithms are Ant Colony Optimization (ACO) (Colorni et al., 1991), Cat Swarm Optimization (CSO) (Chu et al., 2006), Cuckoo Search (CS) (Yang and Deb, 2009), Firefly Algorithm (FA) (Yang, 2010a), Bat Algorithm (BA) (Yang, 2010a), Snake Optimizer (Hashim and Hussien, 2022), Prairie dog optimization algorithm (Ezugwu et al., 2022), Dwarf mongoose optimization algorithm (Agushaka et al., 2022), etc. The socio inspired algorithms that draws inspiration from social behaviors and dynamics observed in nature and society. These algorithms are mainly classified based on social ideologies, sports, social and cultural interaction and colonization. The algorithms associated with these are as follows: Ideology Algorithm (IA) (Teo et al., 2017), Election Algorithm (EA) (Emami and Derakshan, 2015) and Election Campaign Optimization (ECO) (Lv et al., 2010), League Championship Algorithm (LCA) (Kashan, 2009), Teaching Learning Based Optimization (TLBO) (Rao, et al., 2011), Cohort Intelligence (CI) (Kulkarni et al., 2012), Social Learning Optimization (SLO) (Liu et al., 2016) algorithm, human-inspired metaheuristic algorithm (Dehghani et al., 2022), etc. The physics and chemical based algorithms such as Simulated Annealing (Yao, 1995), Harmony Search Algorithms (Geem et al., 2001), Ray Optimization (Kaveh and Khayatazad, 2012), Optical Inspired Algorithm (OIO) (Kashan, 2015), Colliding Bodies Optimization (CBO) (Kaveh and Mahdavi, 2014), RIME Optimization Algorithm (Su et al., 2023), etc. These algorithms have been widely applied to solved the complex problems from different domains such as design engineering, structural engineering, manufacturing problems, combinatorial optimization, power system, clustering, image processing, etc.

The metaheuristic algorithms are stochastic in nature and do not guarantee optimal solutions to all classes of problems. It has been proven in No-Free-Lunch theorem (Wolpert and Macready 1997). It necessitates to explore the search to obtained the better solutions. This motivates to introduce a new nature inspired algorithm referred as Snail Homing and Mating Search (SHMS) algorithm to solve the wide variety of problems. The SHMS algorithm is inspired from the living habitat of snails. The scientific name of the snail is hermaphroditic gastropods. The snails are found in aquatic systems such as lakes, ponds, rivers and oceans, meaning they can live in both marine systems and freshwater. SHSM algorithm exhibits the trail following behavior (Wells and Buckley, 1972; McFaruume, 1980) in the search of mate and while returning home.

Contributions of the proposed work are as follows:

- Introducing a novel Snail Homing and Mating Search (SHMS) algorithm inspired from the living habitat of the snails.
- Several linear and nonlinear benchmark function are solved to investigate the performance of SHMS algorithm and compared with latest metaheuristic algorithms. The SHMS achieved superior results as compared to other algorithms.
- Three cases of the design and economic optimization of shell and tube heat exchanger problem have been solved.

The organization of the paper is as follows: Section 2 describes the natural behavior of snails. The mathematical formulation of SHMS algorithm followed by the flowchart and graphical representation is presented in Section 3. Solution to unimodal (UM) and multimodal (MM) benchmark test function from SHMS algorithm and its statistical comparison along with result discussion are presented in Section 4. Section 5 represents the performance of SHMS algorithm by solving three cases of design and economic optimization shell and tube heat exchanger problem. The conclusions and future recommendations are presented in section 6.

**2. Snail Homing and Mating Search (SHMS) Algorithm**

In this section, the behavior of snails is explained in terms of SHMS algorithm. The mechanism of SHMS algorithm comprises of homing, mate searching, trail following returning to the home. SHMS algorithm is a bio-inspired, population-based optimization method capable of handling complex linear and nonlinear optimization problems across various domains. It leverages principles from nature to iteratively improve a population of potential solutions, ultimately finding the best solution to the given problem.

**2.1 Biological background**

The snails and slugs are basically classified as hermaphroditic gastropods. The class comprises of aquatic, i.e., salt water, fresh water as well as land or terrestrial snails living in humid areas. They have been evolving by continuously adapting to the changing environment. Their basic motivations are satisfying hunger, thirst, finding hiding place or a shelter and reproduce. The snails are in general vegetarian, so to maintain the necessary metabolism and calcium level they keep searching for food (Alfaro, 2007). To facilitate the locomotion associated with it as well as mate-searching they must constantly produce costly mucus which necessarily requires water supply. The mucus lubrication is very necessary for their stomach foot glide across the surfaces. In order to reduce the water evaporation rate and avoid predators, snails generally take shelter in cool and humid places where moisture is available. As the snails are basically trail following gastropods, it is important to mention that the production of mucus trails is the most energy costly component of the snail locomotion (Ng et al., 2013, Hawkins and Hartnoll, 1983). Snails do have eyes however they are not able to see longer neither they can identify the colors (Chernorizov et al. 1994). So, they follow the mucus trail available on the ground and cues (air-borne chemical smell) arising from nearby home/ nest or from the other snails (Arey and Crozier, 1918, 1921). In addition to utilize the mucus trails left behind by every snail to search for food, water, shelter and mates, the trails are also being used for homing, self-organization, identifying conspecifics as well as discriminating between the trails laid by males and females.

**2.2 Homing and Self Organization**

It is referred to as a behavioral pattern of returning of gastropods to the specific resting positions crevices, holes, etc. after feeding excursions (Hawkins and Hartnoll, 1983; Ohgushi, 1954). The homes or shelters are generally of temporal persistence for solitary as well as aggregating collective homers. According to Stephenson (1936) and Cook et al. (1969), several snail species do not necessarily follow same trail route as they left, on the other hand, some of the species even artificially displaced or their trails are washed away they still find the trail route to home. This underscores that the snails follow the trails from the conspecifics as well as prevailing wind help in detecting the air-borne chemical cues guiding towards the resting home McFaruume (1980) and Cook (1979); however, frequent changes in wind direction forces the snails to also resort to trail following.

**2.3 Mate-searching**

It has three major phases, viz. locating of right species, sexual selection and sexual conflict. The mucus trail following is a complementary or alternative mate-searching strategy to air-borne chemical cues (Cook, 1977; Reise, 2007). The snails can identify the conspecific trails especially when are sexually aroused (Townsend, 1974; Nakashima, 1995). The size of the fecund females is generally larger as compared to others. The chances of following such snails are quite high as the fertility chances are much higher. In addition, the trail of infected snail is generally not followed as it may be a sterile snail. According to Lodi and Keone (2016), the mating process sets up a conflict. The sperm donor snail desires to maximize the number of eggs fertilization belonging to the receiver snail. On the other side, the sperm receiver desires to have its eggs fertilized by multiple donors and does not want all the sperms from one donor to reach the eggs. The snails while mating shoot love darts into each other's body. The chemicals contained in the successfully deployed love darts exercises to increase the chances of the donor's sperms to reach the receiver's eggs with reduced chances of wastage, while the receiver's reproductive system has been evolved to counter this donor's manipulation. More specifically, the chemicals contained in the love darts increases the number of contractions between the copulatory canal and sperm digestion sack of the receiver snail, which increases the chances of the donor's sperms to reach the sperm storage sack for egg fertilization. On the other hand, the sperm entry canal referred to as diverticulum tries to increase number of contractions to reduce the number of donor's sperms reaching the copulatory canal. This has in fact given rise to evolution of increasingly potent love-dart mucus.

**3 SHMS Algorithm**

The graphical representation of SHMS algorithm is presented in Figure 1. For the sake of explanation here $H$ ($H = H_1, H_2, H_3$) number of homes are considered. Each home consists of equal number of snails $S$ ($S = S_1, \ldots, S_i, \ldots, S_n$). Every snail $S_i$ initialize the position within the vicinity of their home. Snails continuously travel to find food and a mate, leaving behind a trail of mucus that serves as a guide for their return. Snails tend to navigate by following the available trails on the ground and responding to cues from nearby shelter homes. However, their ability to return to their own homes is limited, resulting in variations in the number of snails in each home. The details mathematical formulation of proposed SHMS algorithm is presented in Section 3.

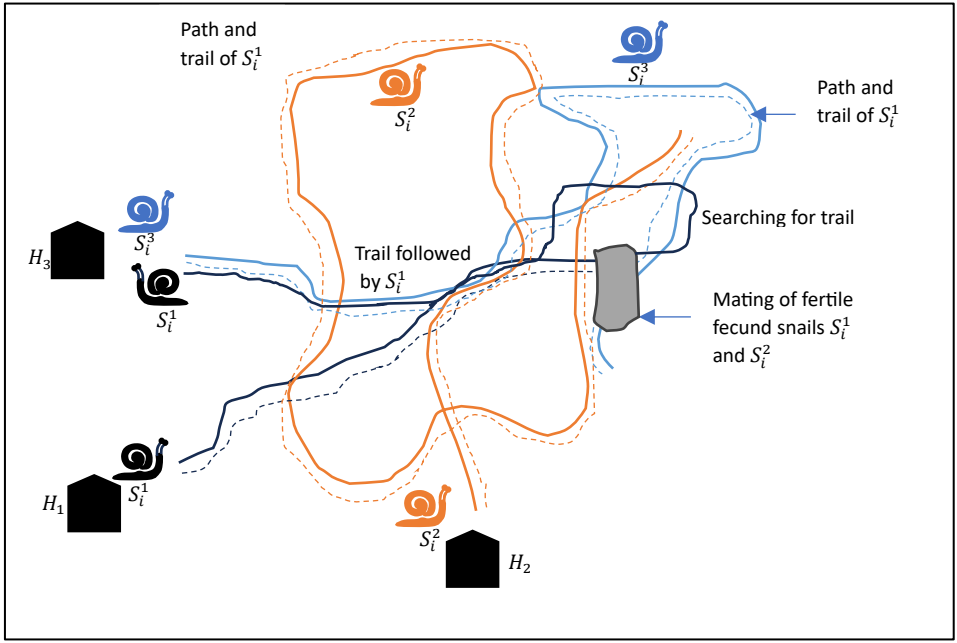

**Fig. 1: Snail path and trail following diagram**

─── indicates the path and ------- indicates the trail

### 3.1 Mathematical Formulation of SHMS Algorithm

Consider the generalized optimization problem in minimization sense as follows:

$$\min f(X) = f(X_1, \ldots, X_i, \ldots, X_n) \quad i = 1,2,3,\ldots,n$$

$$\psi = X_i^l \leq X_i \leq X_i^u$$

**Step 1: Generation of Home and Snails**

Initialize the positions of $H$ homes $X_H = \big((X_1), \ldots, (X_h), \ldots, (X_H)\big)$ and the associated solutions $\big(f(X_1), \ldots, f(X_h), \ldots, f(X_H)\big)$ which are referred to as homes.

Generate $S$ snails at every home $h$ by generating the solution in the close neighborhood of associated home solution $f(X_h)$. The close neighborhood intervals are calculated as follows.

$$\psi_S = X_h \pm c$$

where, $c$ is sampling index which keeps the position of snails within the close neighborhood of its home $f(X_h)$. The sampling space of each snail is calculated only once during the test. Every snail randomly generates the position as follows:

$$X_S^h = \left((X_1), \ldots, (X_s), \ldots, (X_S)\right)^h$$

The snail solutions associated with every home $h, (h = 1, \ldots, H)$ are modeled as follows:

$\left(f(X_1^h), \ldots, f(X_s^h), \ldots, f(X_S^h)\right)$. The matrix represents the arrangement of number of homes and snails.

$$\begin{array}{cccc} \text{Homes} & X_1 & \cdots & X_h & \cdots & X_H \end{array}$$

$$\text{Snails} \begin{bmatrix} (X_1)^1 & (X_1)^h & (X_1)^H \\ \vdots & \vdots & \vdots \\ (X_s)^1 & (X_s)^h & (X_s)^H \\ \vdots & \vdots & \vdots \\ (X_S)^1 & (X_S)^h & (X_S)^H \end{bmatrix}$$

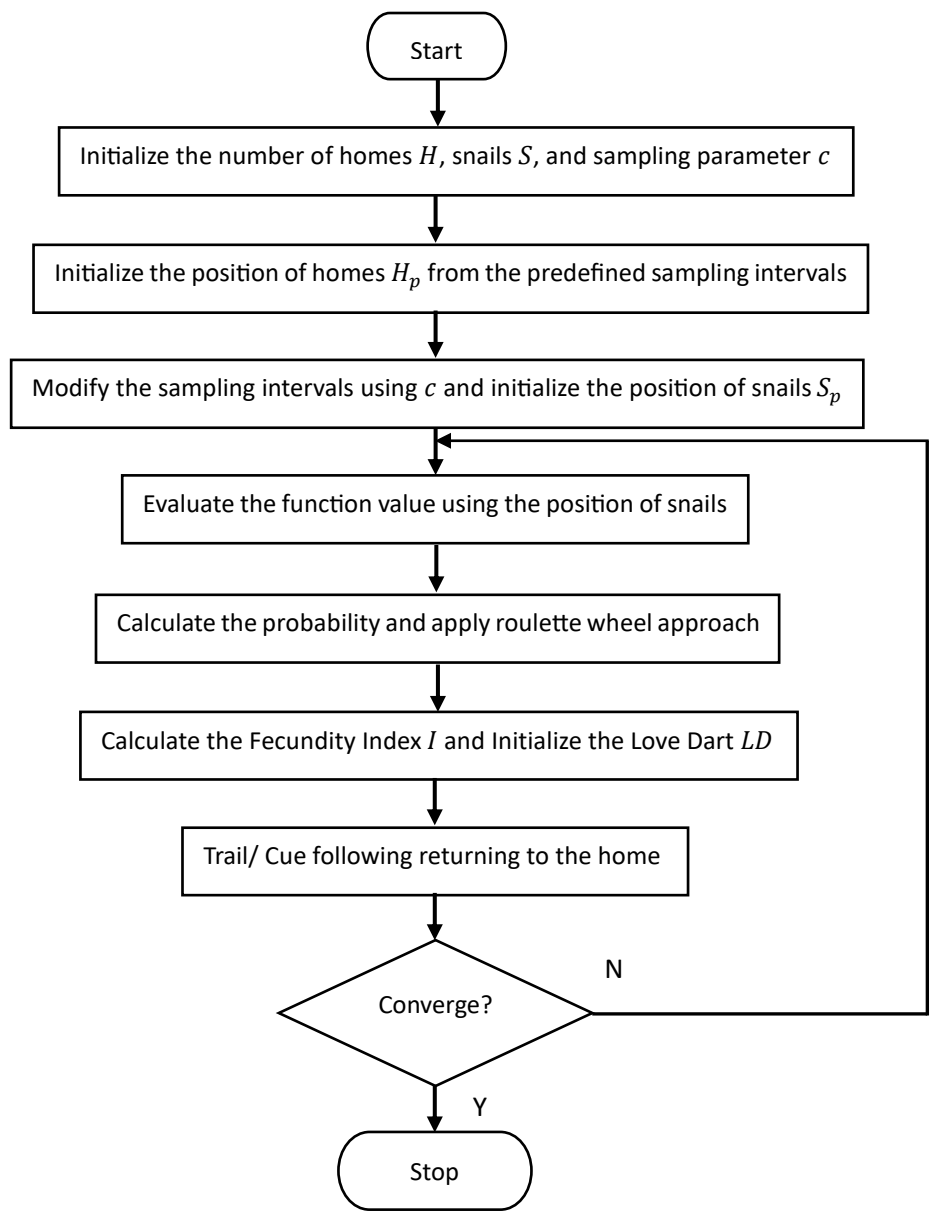

**Fig. 2: Flowchart of SHMS Algorithm**

**Step 2: Calculation of Fecundity Index of snails**

If the solution achieved is better as compared to the earlier fecundity of the snail increases. It is measure using fecundity index as follows:

$$I_{sh} = \left\| \frac{\left(f(X_1^h)^{iter} - f(X_1^h)^{iter-1}\right)}{\left(f(X_1^h)^{iter} - f(X_1^h)^{iter-2}\right)} \right\|.$$

If $I_{sh} \neq 0 \; else \; I_{sh} = rand(0,1)$

The fecundity index $I_{sh}$ increases if the solution consistently improves and decreases if the solution degenerates. The positive value of the fecundity index $I_{sh}$ indicates that the snail is available for mating and negative value indicates that the snail moving towards worsen solution and hence not available for mating.

**Step 3: Calculate the Probability and apply roulette wheel approach**

Calculate the probability of snails

$$P_s = \frac{\left(1/f(X_s^h)\right)}{\sum_{s=1}^{S} \left(1/f(X_s^h)\right)} \quad (s = 1, \ldots, S)$$

Apply the roulette wheel approach to select fecund snail for mating.

**Step 4: Love Dart and Mating**

If a snail consistently exhibits increase in fecundity index, then $s$ snails in certain vicinity using a roulette wheel approach mates with the fecund snail. The love dart $LD$ of the better solution snail are more effective. The $LD$ effect is calculated as follows:

$$LD_s = \frac{1}{I_{sh} \times \left(f(X_s^h) - f\left(X_{s_{fecund}}^h\right)\right)}$$

The better snail solution carries more weightage as compared with other snails.

The value of the $LD_s$ are too large hence it is normalized for computational purpose as follows:

$$LD_s = \frac{(LD_s - (LD_s)_{min})}{(LD_s)_{max} - (LD_s)_{min}}$$

Further the $LD$ effect is utilized to calculate the new sampling interval of the snails.

**Step 5: Trail/ Cue following returning to the home**

The snail having better solution carries more weightage as compared with other snails. The snail solutions with lower probability of obtaining the better solution are biased more towards better solution. Generally, sails follow the trail/ cues while returning to their home. There is a possibility that while returning to their home the snail may follow trail/ cues generated by other fecund snails $s_{fecund}^h$. It may happen the snail $s^h$ can get the home of other snails. According to that, the snail generates a new position in the vicinity of the current home and continue to travel. These sampling intervals associated with every variable are generated as follows:

$$s_{up}^h = LD_s \times \left(s^h - s_{fecund}^h\right)$$

In this paper, several linear and nonlinear test problems and threes cases of shell and tube heat exchanger problem are considered for the validation of proposed SHMS algorithm. The SHMS algorithm is coded in MATLAB R2022a and the simulations are run on Windows platform using 11th Gen Intel(R) Core(TM) i7-1165G7 @ 2.80GHz processor speed and 8 BG RAM. Every individual problem is solved 30 time. The solutions obtained from SHMS algorithm are compared with other well-known nature inspire optimization algorithms discussed in following sections.

**4. Solution to Benchmark test functions**

The SHMS algorithm is validated by solving a set of different benchmark functions from Abdollahzadeh et al 2021. This set of benchmark functions consists of 2 different groups of unimodal (UM) and multi-modal (MM) functions. The function names and features of the UM and MM benchmark function are shown in Table 1- 3. Every problem in these benchmark test cases is solved 30 times using SHMS algorithm.

Table 1 Details of unimodal benchmark functions

| No. | Type | Function | Dimensions | Range | $F_{min}$ |
|---|---|---|---|---|---|
| F1 | US | Sphere | 30, 100, 500, 1000 | [-100, 100]$^d$ | 0 |
| F2 | UN | Schwefel 2.22 | 30, 100, 500, 1000 | [-10, 10]$^d$ | 0 |
| F3 | UN | Schwefel 1.2 | 30, 100, 500, 1000 | [-100, 100]$^d$ | 0 |
| F4 | US | Schwefel 2.21 | 30, 100, 500, 1000 | [-100, 100]$^d$ | 0 |
| F5 | UN | Rosenbrock | 30, 100, 500, 1000 | [-30, 30]$^d$ | 0 |
| F6 | US | Step | 30, 100, 500, 1000 | [-100, 100]$^d$ | 0 |
| F7 | US | Quartic | 30, 100, 500, 1000 | [-128, 128]$^d$ | 0 |

Table 2 Details of multi-modal benchmark functions

| No. | Type | Function | Dimensions | Range | $F_{min}$ |
|---|---|---|---|---|---|
| F8 | MS | Schwefel | 30, 100, 500, 1000 | [-500, 500]$^d$ | −418.9829 × n |
| F9 | MS | Rastrigin | 30, 100, 500, 1000 | [-5.12, 5.12]$^d$ | 0 |
| F10 | MN | Ackley | 30, 100, 500, 1000 | [-32, 32]$^d$ | 0 |
| F11 | MN | Griewank | 30, 100, 500, 1000 | [-600, 600]$^d$ | 0 |
| F12 | MN | Penalized | 30, 100, 500, 1000 | [-50, 50]$^d$ | 0 |
| F13 | MN | Penalized2 | 30, 100, 500, 1000 | [-50, 50]$^d$ | 0 |

Table 3 Details of fixed-dimension multi-modal benchmark functions

| No. | Type | Function | Dimensions | Range | $F_{min}$ |
|---|---|---|---|---|---|
| F14 | FM | Foxholes | 2 | [-65, 65]$^d$ | 1 |
| F15 | FM | Kowalik | 4 | [-5, 5]$^d$ | 0.0003 |
| F16 | FM | Six Hump Camel | 2 | [-5, 5]$^d$ | -1.0316 |
| F17 | FM | Branin | 2 | [-5, 5]$^d$ | 0.398 |
| F18 | FM | Goldstein-Price | 2 | [-2, 2]$^d$ | 3 |
| F19 | FM | Hartman 3 | 3 | [1, 3]$^d$ | -3.86 |
| F20 | FM | Hartman 6 | 6 | [0, 1]$^d$ | -3.32 |
| F21 | FM | Shekel 5 | 4 | [0, 10]$^d$ | -10.1532 |
| F22 | FM | Shekel 7 | 4 | [0, 10]$^d$ | -10.4028 |
| F23 | FM | Shekel 10 | 4 | [0, 10]$^d$ | -10.5363 |

## 4.1 Statistical Analysis

The SHMS algorithm proposed here is compared with several recent algorithms such as African Vultures Optimization Algorithm (AVOA) (Abdollahzadeh et al 2021), PSO (Simon, 2008), Grey Wolf Optimizer (GWO) (Mirjalili et al., 2014), Whale Optimization Algorithm (WOA) (Mirjalili & Lewis, 2016), Farmland Fertility Algorithm (FFA) (Shayanfar & Gharehchopogh, 2018), TLBO (Rao, Savsani, & Vakharia, 2012), Moth-flame optimization algorithm (MFO) (Mirjalili, 2015), Biogeography-Based Optimization (BBO) (Simon, 2008), Differential Evolution (DE) (Simon, 2008), Salp Swarm Algorithm (SSA) (Mirjalili et al., 2017), Gravitational Search Algorithm (GSA) (Rashedi et al., 2009) and Inclined Planes system Optimization (IPO) (Mozaffari et al., 2016). Note that the results for the above-mentioned algorithms were obtained from Abdollahzadeh et al 2021. The comparison of mean solution, best solution and standard deviation (Std. Dev.), mean run time (in seconds) over the 30 runs of the SHMS algorithm solving F1-F13 problems for 30, 100, 500 and 1000 dimensions are represented in Table 4 to Table 7, respectively. The purpose of performing evaluation over different dimensions is to check the Snail algorithms scalability. From the Table 4-7 it is evident that SHMS algorithm demonstrates ability to deliver high quality solutions in various dimensions. In addition, it shows the ability to solve problems having larger dimensions with same set of parameters and without affecting its performance. Table 8 shows the performance comparison of SHMS algorithm for fixed dimension MM problems F14-F23. The algorithm yields comparable results for some of these functions. However, algorithm is not efficient in solving some benchmark functions due to trapping in the local minima. The snails' behavior is visually represented through convergence plots, contour plots, and 3D mesh plots, revealing the outcomes. Figures 3(a)-3(e) represents the performance of SHMS algorithm on UM benchmark test functions. For the MM benchmark test functions, the performance of SHMS algorithm presented in Figures 4(a)-4(i). In these plots the best function values (best snails) are plotted. However, it is also important to showcase the behaviour of all the snails and their homes. The representation of homes and associated snails for two UM and MM are presented in Figure 5. In these plots, the *triangle* refers to the homes and *dots* refer to the associated snails. From the figures it is evident that SHMS algorithm has an ability of quick convergence which obtained the solution with lesser number of iterations. The convergence plot shows the convergence of the solution, contour plot and 3D mesh plot show the search path of the snails within the function.

A pairwise comparison of SHMS algorithm with every other algorithm is carried out, i.e., the values of 30 independent runs solving every problem using SHMS algorithm are compared with every other algorithm solving 30 independent runs of these problems. The Wilcoxon signed-rank test was used for such pairwise comparison with the significance value α chosen to be 0.05. The algorithms with statistically better solutions for F1-F13 problems found using Wilcoxon signed-rank test for 30, 100, 500 and 1000 dimensions are presented in Table 10 to Table 13, respectively. The results highlighted that the SHMS algorithm performed significantly better than every other algorithm for the dimensions 30 and 100 as shown in Table 10 and 11. However, it is evident from the Table 12 and 13 that SHMS algorithm outperformed every other algorithm except AVOA due to the increase in the number of variables increases i.e., 500 and 1000. Table 14 shows the pairwise comparison of F14-F23 problems where SHMS algorithm could not yield better solutions as compared to other algorithms. Results shown in Table 9 demonstrate that the SHMS algorithm is a robust approach with reasonable computational cost and could quickly reach in the close neighbourhood of the global optimum solution. SHMS algorithm not only has found better solutions than other optimization algorithms in the F1-F13 test functions, but it also performs faster than other optimization algorithms in terms of execution time. It even takes less running time to solve large-scale issues and confirms that the computational complexity of SHMS is less than other algorithms. In addition, for statistical analysis Friedman test is performed. The Friedman test is a nonparametric statistical method used to detect multiple test attempts. Using the one-way repeated analysis of variances the ranks were calculated for multiple algorithms by columns. In the Friedman test, the average of optimum solutions obtained from 30 runs solving the F1-F13 problem for 30, 100, 500 and 1000 dimensions and F14–F23 problems using SHMS algorithm is compared with other algorithms solving the same benchmark test functions. The Friedman test ranks are based on the mean values obtained by the algorithms (refer to Tables 15 to 19).

Table 4 Results of benchmark functions (F1-F13), with 30 dimensions

| No. | | AVOA | PSO | GWO | FFA | WOA | TLBO | MFO | BBO | DE | SSA | GSA | IPO | SHMS |
|---|---|---|---|---|---|---|---|---|---|---|---|---|---|---|
| Fl | Best | 5.44E-269 | 1.37E+03 | 6.81E-29 | 2.27E-07 | 2.43E-87 | 1.97E-100 | 1.01E-01 | 3.15E+00 | 7.46E-05 | 2.75E-08 | 2.11E-16 | 1.82E-12 | **0.00E+00** |
| | Worst | 6.05E-198 | 4.27E+03 | 3.14E-26 | 2.17E-06 | 2.01E-69 | 2.26E-95 | 1.00E+04 | 6.33E+00 | 7.89E-04 | 8.75E-07 | 8.65E-03 | 3.88E-10 | **0.00E+00** |
| | Mean | 2.01,199 | 2.37E+03 | 2.59E-27 | 7.88E-07 | 8.04E-71 | 1.66E-96 | 1.66E+03 | 4.47E+00 | 3.36E-04 | 2.07E-07 | 2.88E-04 | 4.76E-11 | **0.00E+00** |
| | STD | 0.00E+00 | 6.17E+02 | 5.83E-27 | 5.15E-07 | 3.72E-70 | 4.36E-96 | 3.78E+03 | 7.54E-01 | 1.83E-04 | 2.17,07 | 1.58E-03 | 8.42E-11 | **0.00E+00** |
| F2 | Best | 2.87E-136 | 1.13E+01 | 1.29E-17 | 1.35E-06 | 5.07E-59 | 1.57E-50 | 8.00E-02 | 3.90E-01 | 1.54E-03 | 2.55E-01 | 5.64E-08 | 6.77E-08 | **0.00E+00** |
| | Worst | 2.58E-102 | 2.77E+01 | 4.34E-16 | 8.00E-06 | 2.16E-48 | 1.11E-48 | 7.00E+01 | 6.00E-01 | 5.95E-03 | 8.00E+00 | 1.19E+00 | 1.27E-05 | **0.00E+00** |
| | Mean | 8.72E-104 | 1.91E+01 | 9.02E-17 | 3.82E-06 | 7.41E-50 | 1.89E-49 | 2.92E+01 | 5.05E-01 | 3.48E-03 | 2.39E+00 | 1.40E-01 | 8.28E-07 | **0.00E+00** |
| | SID | 4.71E-103 | 3.55E+00 | 8.98E-17 | 1.54E-06 | 3.94E-49 | 2.79E-49 | 2.03E+01 | 4.53E-02 | 1.00E-03 | 1.72E+00 | 3.34E-01 | 2.37E-06 | **0.00E+00** |
| F3 | Best | 4.92E-217 | 4.07E+03 | 9.78E-08 | 2.78E+03 | 1.02E+04 | 2.76E-26 | 3.45E+03 | 2.51E+02 | 2.17E+04 | 4.40E+02 | 6.57E+02 | 1.30E+00 | **0.00E+00** |
| | Worst | 2.35E-143 | 2.29E+04 | 7.58E-04 | 7.39E+03 | 6.08E+04 | 2.93E-21 | 4.48E+04 | 1.00E+03 | 5.36E+04 | 5.37E+03 | 1.77E+03 | 4.53E+00 | **0.00E+00** |
| | Mean | 7.83E-145 | 9.20E+03 | 4.02E-05 | 5.03E+03 | 4.18E+04 | 3.35E-22 | 2.23E+04 | 4.98E+02 | 3.64E+04 | 1.83E+03 | 1.09E+03 | 2.41E+00 | **0.00E+00** |
| | STD | 4.29E-144 | 4.24E+03 | 1.46E-04 | 1.32E+03 | 1.32E+04 | 6.76E-22 | 1.19E+04 | 1.49E+02 | 7.01E+03 | 1.02E+03 | 3.11E+02 | 7.86E-01 | **0.00E+00** |
| F4 | Best | 2.57E-134 | 1.57E+01 | 6.93E-08 | 1.33E+01 | 5.27E+00 | 1.02E-42 | 5.39E+01 | 1.21E+00 | 6.27E+00 | 5.22E+00 | 4.27E+00 | 2.41E-02 | **0.00E+00** |
| | Worst | 2.84E-102 | 2.80E+01 | 6.17E-06 | 2.11E+01 | 8.61E+01 | 3.41E-40 | 8.28E+01 | 2.12E+00 | 1.75E+01 | 1.52E+01 | 1.12E+01 | 8.47E-02 | **0.00E+00** |
| | Mean | 1.03E-103 | 2.14E+01 | 7.54E-07 | 1.64E+01 | 4.94E+01 | 9.51E-41 | 6.77E+01 | 1.56E+00 | 9.40E+00 | 1.06E+01 | 7.43E+00 | 4.59E-02 | **0.00E+00** |
| | STD | 5.18E-103 | 3.51E+00 | 1.16E-06 | 2.39E+00 | 2.79E+01 | 1.07E-40 | 8.29E+00 | 2.00E-01 | 2.32E+00 | 2.54E+00 | 1.95E+00 | 1.48E-02 | **0.00E+00** |
| F5 | Best | 1.96E-05 | 1.48E+05 | 2.61E+01 | 1.51E+01 | 2.72E+01 | 2.66E+01 | 1.17E+02 | 4.99E+01 | 2.75E+01 | 2.51E+01 | 2.46E+01 | 1.34E+02 | **2.89E+01** |
| | Worst | 2.56E-02 | 1.62E+06 | 2.88E+01 | 1.85E+02 | 2.84E+01 | 2.84E+01 | 7.99E+07 | 1.64E+03 | 2.38E+02 | 1.08E+03 | 3.27E+02 | 3.96E+02 | **2.90E+01** |
| | Mean | 6.50E-03 | 5.15E+05 | 2.70E+01 | 7.31E+01 | 2.76E+01 | 2.72E+01 | 2.69E+06 | 2.24E+02 | 6.00E+01 | 1.85E+02 | 8.34E+01 | 2.27E+02 | **2.90E+01** |
| | STD | 7.15E-03 | 3.37E+05 | 7.53E-01 | 3.94E+01 | 3.94E-01 | 4.23E-01 | 1.45E+07 | 3.60E+02 | 6.31E+01 | 2.46E+02 | 6.72E+01 | 7.44E+01 | **2.24E-02** |
| F6 | Best | 2.43E-07 | 1.19E+03 | 9.91E-05 | 1.07E-06 | 1.00,01 | 3.37E-07 | 2.80,01 | 1.37E+00 | 7.32E-05 | 8.02,08 | 1.16E-16 | 0.00E+00 | **0.00E+00** |
| | Worst | 6.58E-06 | 3.55E+03 | 1.47E+00 | 1.24E-01 | 8.78E-01 | 5.71E-05 | 1.95E+04 | 3.17E+00 | 5.72E-04 | 6.62E-05 | 5.48E-03 | 4.00E+00 | **0.00E+00** |
| | Mean | 2.43E-06 | 2.35E+03 | 7.34E-01 | 4.26E-03 | 3.80E-01 | 1.06E-05 | 2.66E+03 | 2.36E+00 | 2.65E-04 | 9.76E-06 | 1.83E-04 | 9.33E-01 | **0.00E+00** |
| | STD | 1.89E-06 | 5.34E+02 | 3.06E-01 | 2.25E-02 | 1.91E-01 | 1.28E-05 | 5.18E+03 | 4.18,01 | 1.47E-04 | 1.61E-05 | 1.00E-03 | 1.17E+00 | **0.00E+00** |
| F7 | Best | 4.26E-06 | 1.00,01 | 6.31E-04 | 1.94E-02 | 1.60E-04 | 4.27E-04 | 9.10,02 | 6.34E-03 | 3.00E-02 | 6.10,02 | 2.22E-02 | 9.85E-03 | **3.27E-04** |
| | Worst | 7.61E-04 | 8.91E-01 | 6.15E-03 | 8.19E-02 | 1.46E-02 | 2.66E-03 | 8.25E+00 | 2.67E-02 | 7.44E-02 | 3.17E-01 | 1.60E-01 | 4.36E-02 | **1.62E-02** |
| | Mean | 2.52E-04 | 3.77E-01 | 2.35E-03 | 5.03E-02 | 3.24E-03 | 1.27E-03 | 1.33E+00 | 1.62E-02 | 4.90E-02 | 1.71E-01 | 8.69E-02 | 2.75E-02 | **4.88E-03** |
| | STD | 2.05E-04 | 2.10E-01 | 1.34E-03 | 1.51E-02 | 3.45E-03 | 5.77E-04 | 2.58E+00 | 5.71E-03 | 1.15E-02 | 7.63E-02 | 3.90E-02 | 9.80E-03 | **3.53E-03** |

| No. | | AVOA | PSO | GWO | FFA | WOA | TLBO | MFO | BBO | DE | SSA | GSA | IPO | **SHMS** |
|---|---|---|---|---|---|---|---|---|---|---|---|---|---|---|
| F8 | Best | -1.25E+04 | -5.10E+03 | -7.31E+03 | -9.37E+03 | -1.25E+04 | -9.54E+03 | -1.00E+04 | -9.04E+03 | -7.59E+03 | -9.33E+03 | -3.34E+03 | -4.12E+03 | **0.00E+00** |
| | Worst | -1.21E+04 | -2.69E+03 | -4.69E+03 | -4.88E+03 | -7.98E+03 | -5.86E+03 | -6.47E+03 | -7.01E+03 | -6.05E+03 | -6.14E+03 | -1.70E+03 | -2.52E+03 | **0.00E+00** |
| | Mean | -1.25E+04 | -3.82E+03 | -5.99E+03 | -6.89E+03 | -9.69E+03 | -7.39E+03 | -8.51E+03 | -8.10E+03 | -6.72E+03 | -7.32E+03 | -2.47E+03 | -3.28E+03 | **0.00E+00** |
| | STD | 1.00E+02 | 6.19E+02 | 6.14E+02 | 1.14E+03 | 1.50E+03 | 9.60E+02 | 7.76E+02 | 5.65E+02 | 3.51E+02 | 7.84E+02 | 3.72E+02 | 3.60E+02 | **0.00E+00** |
| F9 | Best | 0.00E+00 | 1.08E+02 | 5.68E-14 | 3.59E+01 | 0.00E+00 | 0.00E+00 | 7.57E+01 | 2.59E+01 | 1.29E+02 | 2.98E+01 | 1.39E+01 | 9.25E+00 | **0.00E+00** |
| | Worst | 0.00E+00 | 2.11E+02 | 1.93E+01 | 1.66E+02 | 0.00E+00 | 4.52E+01 | 2.61E+02 | 8.01E+01 | 1.74E+02 | 9.55E+01 | 4.68E+01 | 3.17E+01 | **0.00E+00** |
| | Mean | 0.00E+00 | 1.48E+02 | 2.00E+00 | 7.96E+01 | 0.00E+00 | 2.08E+01 | 1.55E+02 | 5.22E+01 | 1.57E+02 | 5.05E+01 | 2.92E+01 | 1.68E+01 | **0.00E+00** |
| | STD | 0.00E+00 | 2.16E+01 | 4.07E+00 | 3.10E+01 | 0.00E+00 | 9.68E+00 | 4.15E+01 | 1.30E+01 | 1.06E+01 | 1.58E+01 | 7.25E+00 | 4.32E+00 | **0.00E+00** |
| F10 | Best | 8.88E-16 | 6.71E+00 | 7.19E-14 | 9.29E-05 | 8.84E-16 | 4.44E-15 | 5.58E-01 | 3.80E-01 | 3.04E-03 | 1.34E+00 | 7.68E-09 | 1.61E+00 | **4.44E-16** |
| | Worst | 8.88E-16 | 1.20E+01 | 1.35E-13 | 8.73E-04 | 7.95E-15 | 7.99E-15 | 2.00E+01 | 7.63E-01 | 7.78E-03 | 6.28E+00 | 1.16E+00 | 3.83E+00 | **4.44E-16** |
| | Mean | 8.88E-16 | 1.02E+01 | 9.70E-14 | 3.44E-04 | 4.08E-15 | 5.98E-15 | 1.40E+01 | 5.90E-01 | 5.19E-03 | 2.99E+00 | 6.95E-02 | 2.30E+00 | **4.44E-16** |
| | STD | 0.00E+00 | 1.30E+00 | 1.55E-14 | 1.62E-04 | 2.35E-15 | 1.79E-15 | 7.67E+00 | 9.94E-02 | 1.32E-03 | 1.19E+00 | 2.66E-01 | 4.82E-01 | **0.00E+00** |
| F11 | Best | 0.00E+00 | 1.22E+01 | 0.00E+00 | 2.77E-06 | 0.00E+00 | 0.00E+00 | 6.32E-01 | 9.32E-01 | 3.26E-04 | 1.59E-03 | 1.21E+01 | 1.02E-03 | **0.00E+00** |
| | Worst | 0.00E+00 | 3.24E+01 | 2.93E-02 | 3.53E-02 | 1.89E-01 | 1.20E-04 | 9.10E+01 | 1.07E+00 | 1.29E-01 | 6.12E-02 | 4.75E+01 | 4.30E-02 | **0.00E+00** |
| | Mean | 0.00E+00 | 2.11E+01 | 4.70E-03 | 6.41E-03 | 2.21E-02 | 4.98E-06 | 8.37E+00 | 1.01E+00 | 1.44E-02 | 1.69E-02 | 2.88E+01 | 1.10E-02 | **0.00E+00** |
| | STD | 0.00E+00 | 5.28E+00 | 8.48E-03 | 7.85E-03 | 5.38E-02 | 2.23E-05 | 2.34E+01 | 2.46E-02 | 3.48E-02 | 1.26E-02 | 6.76E+00 | 9.04E-03 | **0.00E+00** |
| F12 | Best | 1.72E-08 | 9.28E+00 | 1.26E-02 | 7.85E-05 | 5.89E-03 | 6.03E-10 | 2.08E+00 | 2.94E-03 | 5.02E-05 | 1.44E+00 | 1.04E-01 | 3.05E-02 | **0.00E+00** |
| | Worst | 1.29E-06 | 3.58E+03 | 1.15E-01 | 1.96E+00 | 3.17E-01 | 1.81E-05 | 5.04E+03 | 1.38E-02 | 1.46E-03 | 1.43E+01 | 4.34E+00 | 1.46E+00 | **0.00E+00** |
| | Mean | 3.79E-07 | 2.74E+02 | 5.22E-02 | 1.32E-01 | 2.95E-02 | 1.25E-06 | 1.80E+02 | 8.11E-03 | 3.92E-04 | 7.24E+00 | 2.01E+00 | 3.75E-01 | **0.00E+00** |
| | STD | 3.32E-07 | 6.87E+02 | 2.56E-02 | 3.72E-01 | 5.49E-02 | 3.53E-06 | 9.18E+02 | 2.47E-03 | 4.13E-04 | 3.26E+00 | 1.13E+00 | 3.69E-01 | **0.00E+00** |
| F13 | Best | 8.67E-07 | 1.36E+04 | 6.35E-02 | 1.01E-04 | 1.44E-01 | 4.00E-02 | 4.14E+00 | 8.53E-02 | 2.20E-04 | 2.09E-02 | 4.00E-02 | 6.38E-02 | **1.74E+00** |
| | Worst | 8.85E-05 | 9.43E+05 | 1.05E+00 | 2.89E-01 | 1.37E+00 | 1.01E+00 | 1.91E+03 | 1.80E-01 | 7.14E-03 | 4.88E+01 | 3.69E+01 | 2.71E-01 | **3.00E+00** |
| | Mean | 1.10E-05 | 1.66E+05 | 6.10E-01 | 4.78E-02 | 5.38E-01 | 4.38E-01 | 1.24E+02 | 1.23E-01 | 1.60E-03 | 1.80E+01 | 1.17E+01 | 1.21E-01 | **2.76E+00** |
| | STD | 1.56E-05 | 2.15E+05 | 2.63E-01 | 7.61E-02 | 3.19E-01 | 2.69E-01 | 3.82E+02 | 2.52E-02 | 1.59E-03 | 1.41E+01 | 8.46E+00 | 4.09E-02 | **3.11E-01** |

Table 5 Results of benchmark functions (F1-F13), with 100 dimensions

| No. | | AVOA | PSO | GWO | FFA | WOA | TLBO | MFO | BBO | DE | SSA | GSA | IPO | **SHMS** |
|---|---|---|---|---|---|---|---|---|---|---|---|---|---|---|
| F1 | Best | 4.37E-260 | 9.85E+03 | 3.22E-13 | 1.75E+03 | 7.24E-84 | 2.00E-92 | 3.00E+04 | 1.66E+02 | 2.44E+03 | 7.65E+02 | 2.74E+03 | 3.59E-01 | **0.00E+00** |
| | Worst | 2.77E-193 | 2.44E+04 | 8.28E-12 | 3.29E+03 | 1.65E-69 | 1.68E-89 | 8.90E+04 | 2.64E+02 | 5.31E+03 | 2.73E+03 | 6.84E+03 | 2.72E+00 | **0.00E+00** |
| | Mean | 1.35E-194 | 1.38E+04 | 1.89E-12 | 2.43E+03 | 5.62E-71 | 1.30E-90 | 5.84E+04 | 2.17E+02 | 3.50E+03 | 1.51E+03 | 4.25E+03 | 1.01E+00 | **0.00E+00** |
| | STD | 0.00E+00 | 3.24E+03 | 1.69E-12 | 4.00E+02 | 3.02E-70 | 3.09E-90 | 1.30E+04 | 2.26E+01 | 7.32E+02 | 4.62E+02 | 8.52E+02 | 6.69E-01 | **0.00E+00** |
| F2 | Best | 1.75E-139 | 6.96E+01 | 1.31E-08 | 1.72E+01 | 2.78E-56 | 2.55E-47 | 1.95E+02 | 9.57E+00 | 3.40E+01 | 2.94E+01 | 1.26E+01 | 4.08E+00 | **0.00E+00** |
| | Worst | 4.65E-103 | 1.06E+02 | 5.88E-08 | 5.20E+01 | 1.02E-48 | 1.48E-45 | 3.39E+02 | 1.26E+01 | 8.40E+01 | 1.25E+02 | 3.38E+01 | 1.01E+01 | **0.00E+00** |
| | Mean | 1.66E-104 | 8.92E+01 | 3.85E-08 | 2.65E+01 | 6.24E-50 | 4.27E-46 | 2.42E+02 | 1.01E+01 | 6.26E+01 | 5.13E+01 | 1.84E+01 | 6.53E+00 | **0.00E+00** |
| | STD | 8.49E-104 | 9.79E+00 | 1.07E-08 | 7.18E+00 | 2.02E-49 | 4.36E-46 | 2.80E+01 | 7.06E-01 | 1.20E+01 | 1.63E+01 | 4.78E+00 | 1.48E+00 | **0.00E+00** |
| F3 | Best | 2.06E-208 | 2.98E+04 | 3.91E+01 | 1.54E+05 | 6.46E+05 | 5.39E-14 | 1.17E+05 | 3.90E+04 | 3.04E+05 | 1.65E+04 | 8.84E+03 | 2.90E+03 | **0.00E+00** |
| | Worst | 3.71E-132 | 1.86E+05 | 2.93E+03 | 2.33E+05 | 1.71E+06 | 1.16E-08 | 3.35E+05 | 7.08E+04 | 5.73E+05 | 1.01 e+05 | 2.64E+04 | 9.60E+03 | **0.00E+00** |
| | Mean | 1.28E-133 | 9.60E+04 | 6.03E+02 | 2.14E+05 | 1.06E+06 | 7.90E-10 | 2.34E+05 | 5.01E+04 | 4.76E+05 | 5.10E+04 | 1.54E+04 | 4.76E+03 | **0.00E+00** |
| | STD | 6.77E-133 | 3.64E+04 | 6.81E+02 | 1.56E+04 | 2.92E+05 | 2.51E-09 | 6.22E+04 | 8.35E+03 | 6.00E+04 | 2.45E+04 | 4.59E+03 | 1.30E+03 | **0.00E+00** |
| F4 | Best | 2.62E-129 | 2.41E+01 | 1.49E-01 | 8.88E+01 | 2.74E+01 | 7.36E-39 | 8.77E+01 | 1.54E+01 | 9.02E+01 | 2.30E+01 | 1.63E+01 | 8.64E+00 | **0.00E+00** |
| | Worst | 2.74E-100 | 4.00E+01 | 2.37E+00 | 9.76E+01 | 9.77E+01 | 4.44E-37 | 9.74E+01 | 2.75E+01 | 9.80E+01 | 3.75E+01 | 2.32E+01 | 1.37E+01 | **0.00E+00** |
| | Mean | 9.17E-102 | 3.07E+01 | 5.79E-01 | 9.53E+01 | 8.22E+01 | 1.17e-.37 | 9.36E+01 | 2.04E+01 | 9.49E+01 | 2.92E+01 | 1.95E+01 | 1.06E+01 | **0.00E+00** |
| | STD | 5.00E-101 | 3.75E+00 | 4.84E-01 | 1.81E+00 | 1.67E+01 | 1.09E-37 | 2.11E+00 | 2.93E+00 | 1.85E+00 | 3.46E+00 | 1.90E+00 | 1.30E+00 | **0.00E+00** |
| F5 | Best | 4.67E-04 | 2.44E+06 | 9.59E+01 | 4.28E+06 | 9.77E+01 | 9.68E+01 | 6.30E+07 | 4.19E+03 | 2.42E+06 | 4.71E+04 | 3.66E+04 | 5.28E+03 | **2.89E+01** |
| | Worst | 2.88E-01 | 7.32E+06 | 9.81E+01 | 2.10E+07 | 9.81E+01 | 9.84E+01 | 2.56E+08 | 7.36E+03 | 8.88E+06 | 2.55E+05 | 2.63E+05 | 2.59E+04 | **2.90E+01** |
| | Mean | 5.12E-02 | 4.75E+06 | 9.77E+01 | 1.17E+07 | 9.79E+01 | 9.76E+01 | 1.50E+08 | 5.31E+03 | 5.34E+06 | 1.31E+05 | 1.15E+05 | 1.07E+04 | **2.90E+01** |
| | STD | 7.27E-02 | 1.34E+06 | 5.87E-01 | 3.66E+06 | 2.15E-01 | 3.78E-01 | 5.84E+07 | 9.20E+02 | 1.61E+06 | 5.60E+04 | 6.87E+04 | 3.85E+03 | **1.64E-02** |
| F6 | Best | 3.29E-06 | 1.04E+04 | 6.88E+00 | 1.36E+03 | 2.10E+00 | 4.72E+00 | 2.79E+04 | 1.76E+02 | 2.38E+03 | 6.53E+02 | 2.96E+03 | 3.40E+01 | **0.00E+00** |
| | Worst | 3.51E-03 | 1.89E+04 | 1.32E+01 | 3.29E+03 | 6.92E+00 | 9.46E+00 | 1.00E+05 | 2.99E+02 | 4.81E+03 | 2.72E+03 | 8.75E+03 | 1.40E+03 | **0.00E+00** |
| | Mean | 7.23E-04 | 1.44E+04 | 1.00E+01 | 2.41E+03 | 4.26E+00 | 7.39E+00 | 5.91E+04 | 2.30E+02 | 3.39E+03 | 1.52E+03 | 4.65E+03 | 1.27E+02 | **0.00E+00** |
| | STD | 1.00E-03 | 2.38E+03 | 1.55E+00 | 4.63E+02 | 1.25E+00 | 1.08E+00 | 1.45E+04 | 2.57E+01 | 6.31E+02 | 4.43E+02 | 1.10E+03 | 2.50E+02 | **0.00E+00** |
| F7 | Best | 9.38E-06 | 3.15E+00 | 2.54E-03 | 9.39E+00 | 5.81E-05 | 5.17E-04 | 5.24E+01 | 8.27E-02 | 3.39E+00 | 1.50E+00 | 2.15E+00 | 9.33E-01 | **9.29E-05** |
| | Worst | 4.68E-04 | 1.34E+01 | 1.08E-02 | 2.06E+01 | 1.27E-02 | 3.20E-03 | 5.24E+02 | 1.68E-01 | 1.21E+01 | 4.72E+00 | 9.99E+00 | 3.60E+01 | **2.53E-02** |
| | Mean | 1.83E-04 | 7.55E+00 | 6.72E-03 | 1.37E+01 | 3.53E-03 | 1.77E-03 | 2.41E+02 | 1.25E-01 | 6.56E+00 | 2.75E+00 | 4.38E+00 | 4.49E+00 | **6.30E-03** |
| | STD | 1.38E-04 | 2.64E+00 | 2.26E-03 | 3.33E+00 | 3.52E-03 | 6.45E-04 | 1.11E+02 | 2.63E-02 | 2.14E+00 | 7.48E-01 | 1.82E+00 | 7.89E+00 | **5.78E-03** |

| No. | | AVOA | PSO | GWO | FFA | WOA | TLBO | MFO | BBO | DE | SSA | GSA | IPO | **SHMS** |
|---|---|---|---|---|---|---|---|---|---|---|---|---|---|---|
| F8 | Best | -4.15E+04 | -1.06E+04 | -2.01 e+04 | -1.92E+04 | -4.18E+04 | -2.34E+04 | -2.58E+04 | -2.55E+04 | -1.37E+04 | -2.59E+04 | -6.02E+03 | -1.36E+04 | **-4.84E+03** |
| | Worst | -4.12e-F04 | -6.28E+03 | -5.85E+03 | -8.89E+03 | -2.52E+04 | -1.05E+04 | -1.82E+04 | -2.02E+04 | -1.09E+04 | -1.79E+04 | -2.87E+03 | -4.91E+03 | **-2.24E+01** |
| | Mean | -4.14E+04 | -7.62E+03 | -1.58E+04 | -1.36E+04 | -3.40E+04 | -1.69E+04 | -2.17E+04 | -2.25E+04 | -1.18E+04 | -2.14E+04 | -4.05E+03 | -1.07E+04 | **-1.23E+03** |
| | STD | 5.28E+01 | 1.13E+03 | 3.05E+03 | 3.09E+03 | 5.69E+03 | 2.71E+03 | 1.83E+03 | 1.03E+03 | 6.65E+02 | 1.73E+03 | 7.99E+02 | 1.93E+03 | **1.09E+03** |
| F9 | Best | 0.00E+00 | 6.66E+02 | 1.38E-10 | 6.44E+02 | 0.00E+00 | 0.00E+00 | 7.43E+02 | 2.55E+02 | 9.10E+02 | 1.54E+02 | 1.37E+02 | 2.40E+02 | **0.00E+00** |
| | Worst | 0.00E+00 | 8.35E+02 | 2.39E+01 | 1.08E+03 | 0.00E+00 | 0.00E+00 | 1.00E+03 | 3.98E+02 | 1.02E+03 | 3.03E+02 | 2.65E+02 | 3.43E+02 | **0.00E+00** |
| | Mean | 0.00E+00 | 7.36E+02 | 8.83E+00 | 8.92E+02 | 0.00E+00 | 0.00E+00 | 8.64E+02 | 3.17E+02 | 9.80E+02 | 2.35E+02 | 1.93E+02 | 2.79E+02 | **0.00E+00** |
| | STD | 0.00E+00 | 4.13E+01 | 6.15E+00 | 1.19E+02 | 0.00E+00 | 0.00E+00 | 7.54E+01 | 3.33E+01 | 2.85E+01 | 3.87E+01 | 3.64E+01 | 2.58E+01 | **0.00E+00** |
| F10 | Best | 8.88E-16 | 1.09E+01 | 7.13E-08 | 7.59E+00 | 8.88E-16 | 4.44E-15 | 1.94E+01 | 3.18E+00 | 7.87E+00 | 8.38E+00 | 3.54E+00 | 4.27E+00 | **4.44E-16** |
| | Worst | 8.88E-16 | 1.32E+01 | 3.21E-07 | 9.84E+00 | 7.99E-15 | 7.99E-15 | 2.03E+01 | 3.56E+00 | 1.00E+01 | 1.23E+01 | 6.83E+00 | 6.45E+00 | **4.44E-16** |
| | Mean | 8.88E-16 | 1.19E+01 | 1.30E-07 | 8.68E+00 | 4.08E-15 | 7.63E-15 | 1.99E+01 | 3.42E+00 | 9.11E+00 | 1.01E+01 | 4.96E+00 | 4.93E+00 | **4.44E-16** |
| | STD | 0.00E+00 | 6.76E-01 | 6.01E-08 | 5.81E-01 | 1.94E-15 | 1.05E-15 | 1.31E-01 | 1.19E-01 | 6.18E-01 | 1.10E+00 | 7.62E-01 | 5.55E-01 | **0.00E+00** |
| F11 | Best | 0.00E+00 | 9.56E+01 | 1.93E-13 | 1.48E+01 | 0.00E+00 | 0.00E+00 | 3.55E+02 | 2.65E+00 | 1.64E+01 | 6.66E+00 | 6.10E+02 | 3.47E-01 | **0.00E+00** |
| | Worst | 0.00E+00 | 1.77E+02 | 3.18E-02 | 3.22E+01 | 0.00E+00 | 0.00E+00 | 7.09E+02 | 3.59E+00 | 5.61E+01 | 2.20E+01 | 7.86E+02 | 1.09E+00 | **0.00E+00** |
| | Mean | 0.00E+00 | 1.23E+02 | 5.30E-03 | 2.29E+01 | 0.00E+00 | 0.00E+00 | 5.33E+02 | 3.17E+00 | 3.13E+01 | 1.34E+01 | 6.92E+02 | 8.22E-01 | **0.00E+00** |
| | STD | 0.00E+00 | 2.15E+01 | 1.09E-02 | 4.15E+00 | 0.00E+00 | 0.00E+00 | 1.05E+02 | 2.01E-01 | 8.05E+00 | 3.60E+00 | 4.12E+01 | 2.48E-01 | **0.00E+00** |
| F12 | Best | 4.01E-07 | 4.25E+02 | 1.46E-01 | 1.09E+07 | 1.83E-02 | 7.43E-02 | 5.11E+07 | 1.22E+00 | 3.46E+06 | 1.20E+01 | 4.70E+00 | 5.24E+00 | **0.00E+00** |
| | Worst | 6.36E-05 | 6.15E+05 | 5.08E-01 | 5.79E+07 | 1.30E-01 | 1.62E-01 | 6.24E+08 | 1.27E+01 | 1.94E+07 | 5.70E+01 | 1.93E+01 | 8.15E+00 | **0.00E+00** |
| | Mean | 6.55E-06 | 1.21E+05 | 3.07E-01 | 2.26E+07 | 5.34E-02 | 1.17E-01 | 2.84E+08 | 4.05E+00 | 9.18E+06 | 3.23E+01 | 1.17E+01 | 6.92E+00 | **0.00E+00** |
| | STD | 1.25E-05 | 1.30E+05 | 7.64E-02 | 1.01E+07 | 2.89E-02 | 2.31E-02 | 1.62E+08 | 2.41E+00 | 4.31E+06 | 9.77E+00 | 3.91E+00 | 8.03E-01 | **0.00E+00** |
| F13 | Best | 3.39E-05 | 7.78E+05 | 5.76E+00 | 1.74E+07 | 1.26E+00 | 6.62E+00 | 1.76E+08 | 9.06E+00 | 6.15E+06 | 1.60E+02 | 1.86E+02 | 5.39E+00 | **9.99E+00** |
| | Worst | 2.88E-03 | 8.58E+06 | 7.51E+00 | 8.77E+07 | 4.62E+00 | 9.92E+00 | 1.14E+09 | 1.34E+01 | 3.56E+07 | 5.85E+04 | 3.02E+04 | 1.60E+02 | **1.00E+01** |
| | Mean | 7.54E-04 | 3.39E+06 | 6.72E+00 | 5.33E+07 | 2.75E+00 | 8.09E+00 | 5.71E+08 | 1.13E+01 | 1.68E+07 | 6.04E+03 | 4.90E+03 | 3.59E+01 | **1.00E+01** |
| | STD | 7.67E-04 | 2.04E+06 | 3.97E-01 | 1.87E+07 | 9.29E-01 | 9.62E-01 | 2.90E+08 | 1.18E+00 | 7.34E+06 | 1.19E+04 | 7.72E+03 | 4.23E+01 | **2.91E-03** |

Table 6 Results of benchmark functions (F1-F13), with 500 dimensions

| No. | | AVOA | PSO | GWO | FFA | WOA | TLBO | MFO | BBO | DE | SSA | GSA | IPO | SHMS |
|---|---|---|---|---|---|---|---|---|---|---|---|---|---|---|
| F1 | Best | 3.18E-249 | 7.46E+04 | 6.20E-04 | 4.86E+05 | 1.40E-83 | 3.47E-88 | 1.05E+06 | 6.32E+03 | 4.89E+05 | 7.64E+04 | 5.02E+04 | 9.28E+03 | **0.00E+00** |
| | Worst | 2.94E-199 | 1.19E+05 | 2.89E-03 | 5.62E+05 | 6.09E-66 | 4.43E-85 | 1.22E+06 | 8.16E+03 | 6.17E+05 | 1.05E+05 | 6.22E+04 | 1.39E+04 | **0.00E+00** |
| | Mean | 1.04E-200 | 9.80E+04 | 1.57E-03 | 5.25E+05 | 2.05E-67 | 4.13E-86 | 1.12E+06 | 7.15E+03 | 5.63E+05 | 9.33E+04 | 5.57E+04 | 1.13E+04 | **0.00E+00** |
| | STD | 0.00E+00 | 1.30E+04 | 5.47E-04 | 2.02E+04 | 1.11,66 | 9.06E-86 | 3.11E+04 | 5.43E+02 | 2.92E+04 | 8.25E+03 | 2.82E+03 | 1.17E+03 | **0.00E+00** |
| F2 | Best | 2.15E-131 | 4.49E+02 | 8.12E-03 | 1.77E+52 | 2.13E-55 | 4.25E-45 | 5.79E+75 | 2.05E+02 | 1.44E+03 | 4.97E+02 | 2.80E+02 | 1.64E+02 | **0.00E+00** |
| | Worst | 1.00E-99 | 6.36E+02 | 1.42E-02 | 7.09E+117 | 9.69E-47 | 3.44E-43 | 1.77E+120 | 2.75E+02 | 1.54E+03 | 5.75E+02 | 3.23E+270 | 1.98E+02 | **0.00E+00** |
| | Mean | 3.38E-101 | 5.33E+02 | 1.09E-02 | 2.36E+116 | 3.89E-48 | 3.95E-44 | 6.04E+118 | 2.29E+02 | 1.50E+03 | 5.31E+02 | 1.08E+269 | 1.81E+02 | **0.00E+00** |
| | STD | 1.84E-100 | 4.81E+01 | 1.58E-03 | 1.29E+117 | 1.78E-47 | 6.78E-44 | 3.24E+119 | 1.72E+01 | 3.50E+01 | 2.00E+01 | 1.00E+300 | 9.43E+00 | **0.00E+00** |
| F3 | Best | 7.15E-196 | 1.00E+06 | 2.33E+05 | 4.74E+06 | 1.49E+07 | 6.47E-08 | 3.31E+06 | 1.19E+06 | 8.87E+06 | 5.20E+05 | 3.92E+05 | 1.10E+05 | **0.00E+00** |
| | Worst | 8.95E-102 | 5.44E+06 | 5.27E+05 | 6.41E+06 | 5.22E+07 | 8.04E-03 | 6.39E+06 | 2.34E+06 | 1.32E+07 | 2.39E+06 | 3.88E+06 | 2.20E+05 | **0.00E+00** |
| | Mean | 2.98E-103 | 2.29E+06 | 3.57E+05 | 5.57E+06 | 3.04E+07 | 6.76E-04 | 4.80E+06 | 1.55E+06 | 1.16E+07 | 1.24E+06 | 1.17E+06 | 1.48E+05 | **0.00E+00** |
| | STD | 1.63E-102 | 8.31E+05 | 7.70E+04 | 4.80E+05 | 1.00E+07 | 1.99E-03 | 9.16E+05 | 2.25E+05 | 1.10E+06 | 6.10E+05 | 6.80E+05 | 2.78E+04 | **0.00E+00** |
| F4 | Best | 1.87E-136 | 3.44E+01 | 5.69E+01 | 9.83E+01 | 1.62E+01 | 8.57E-37 | 9.82E+01 | 4.86E+01 | 9.88E+01 | 3.50E+01 | 2.55E+01 | 1.84E+01 | **0.00E+00** |
| | Worst | 3.49E-100 | 4.77E+01 | 7.77E+01 | 9.92E+01 | 9.87E+01 | 4.51E-35 | 9.91E+01 | 5.75E+01 | 9.92E+01 | 4.86E+01 | 3.27E+01 | 2.19E+01 | **0.00E+00** |
| | Mean | 1.47E-101 | 3.82E+01 | 6.47E+01 | 9.89E+01 | 8.19E+01 | 8.65e-.36 | 9.85E+01 | 5.27E+01 | 9.89E+01 | 4.03E+01 | 2.87E+01 | 2.04E+01 | **0.00E+00** |
| | STD | 6.43E-101 | 2.73E+00 | 5.77E+00 | 3.01E-01 | 2.10E+01 | 9.83E-36 | 3.71E-01 | 2.07E+00 | 1.98E-01 | 3.30E+00 | 1.56E+00 | 9.22E-01 | **0.00E+00** |
| F5 | Best | 3.34E-03 | 2.16E+07 | 4.93E+02 | 1.72E+09 | 4.91E+02 | 4.94E+02 | 4.64E+09 | 6.08E+05 | 1.75E+09 | 2.75E+07 | 6.49E+06 | 2.31E+06 | **2.89E+01** |
| | Worst | 1.86E+01 | 6.50E+07 | 4.96E+02 | 2.95E+09 | 4.97E+02 | 4.97E+02 | 5.45E+09 | 1.01E+06 | 6.00E+09 | 4.72E+07 | 1.40E+07 | 3.53E+06 | **2.90E+01** |
| | Mean | 3.66E+00 | 4.31E+07 | 4.94E+02 | 2.30E+09 | 4.93E+02 | 4.95E+02 | 5.02E+09 | 8.44E+05 | 2.82E+09 | 3.71E+07 | 8.69E+06 | 2.85E+06 | **2.90E+01** |
| | STD | 4.12E+00 | 1.13E+07 | 3.30E-01 | 3.17E+08 | 3.52E-01 | 1.13E-01 | 2.21E+08 | 9.37E+04 | 8.04E+08 | 4.14E+06 | 1.62E+06 | 3.81E+05 | **1.83E-02** |
| F6 | Best | 2.76E-04 | 7.35E+04 | 8.75E+01 | 4.72E+05 | 1.89E+01 | 9.06E+01 | 1.05E+06 | 6.66E+03 | 4.87E+05 | 7.57E+04 | 5.00E+04 | 1.36E+04 | **0.00E+00** |
| | Worst | 4.78E-01 | 1.20E+05 | 9.44E+01 | 5.73E+05 | 4.73E+01 | 9.82E+01 | 1.21E+06 | 8.51E+03 | 6.23E+05 | 1.08E+05 | 6.27E+04 | 2.71E+04 | **0.00E+00** |
| | Mean | 5.90E-02 | 9.65E+04 | 9.14E+01 | 5.29E+05 | 3.24E+01 | 9.42E+01 | 1.15E+06 | 7.35E+03 | 5.50E+05 | 9.42E+04 | 5.70E+04 | 1.96E+04 | **0.00E+00** |
| | STD | 1.03E-01 | 1.42E+04 | 1.68E+00 | 2.55E+04 | 8.30E+00 | 2.05E+00 | 3.50E+04 | 4.77E+02 | 3.53E+04 | 6.51E+03 | 2.75E+03 | 4.06E+03 | **0.00E+00** |
| F7 | Best | 2.71E-06 | 1.74E+02 | 2.62E-02 | 1.23E+04 | 1.43E-04 | 5.14E-04 | 3.28E+04 | 3.67E+02 | 1.16E+04 | 2.08E+02 | 7.08E+02 | 1.77E+03 | **2.60E-03** |
| | Worst | 5.53E-04 | 6.70E+02 | 7.89E-02 | 1.91E+04 | 1.40E-02 | 3.30E-03 | 4.23E+04 | 7.03E+02 | 2.04E+04 | 3.54E+02 | 1.44E+03 | 4.79E+03 | **2.84E-02** |
| | Mean | 2.09E-04 | 3.50E+02 | 5.20E-02 | 1.59E+04 | 4.65E-03 | 1.67E-03 | 3.87E+04 | 4.96E+02 | 1.55E+04 | 2.75E+02 | 9.88E+02 | 2.86E+03 | **1.02E-02** |
| | STD | 1.54E-04 | 1.35E+02 | 1.36E-02 | 1.49E+03 | 4.70E-03 | 6.18E-04 | 2.31E+03 | 9.00E+01 | 2.19E+03 | 3.96E+01 | 1.74E+02 | 7.30E+02 | **6.91E-03** |

| No. | | AVOA | PSO | GWO | FFA | WOA | TLBO | MFO | BBO | DE | SSA | GSA | IPO | SHMS |
|---|---|---|---|---|---|---|---|---|---|---|---|---|---|---|
| F8 | Best | -2.13e-F05 | -2.25E+04 | -6.50E+04 | -3.92E+04 | -2.09E+05 | -6.07E+04 | -7.14E+04 | -7.49E+04 | -2.76E+04 | -6.80E+04 | -1.61E+04 | -4.85E+04 | **-7.90E+03** |
| | Worst | -2.10e-F05 | -1.34E+04 | -4.96E+04 | -1.84E+04 | -1.26E+05 | -2.57E+04 | -5.01E+04 | -6.62E+04 | -2.38E+04 | -4.93E+04 | -6.32E+03 | -9.29E+03 | **-5.07E+01** |
| | Mean | -2.12E+05 | -1.78E+04 | -5.80E+04 | -2.73E+04 | -1.69E+05 | -3.96E+04 | -6.22E+04 | -7.07E+04 | -2.55E+04 | -6.08E+04 | -1.10E+04 | -3.06E+04 | **-2.75E+03** |
| | STD | 6.16E+02 | 2.57E+03 | 3.75E+03 | 5.92E+03 | 3.09E+04 | 9.39E+03 | 4.89E+03 | 2.53E+03 | 1.12E+03 | 4.37E+03 | 2.27E+03 | 1.44E+04 | **2.76E+03** |
| F9 | Best | 0.00e-F00 | 4.28E+03 | 2.37E+01 | 6.50E+03 | 0.00E+00 | 0.00E+00 | 6.61E+03 | 5.75E+03 | 6.51E+03 | 2.95E+03 | 2.53E+03 | 3.17E+03 | **0.00E+00** |
| | Worst | 0.00e-F00 | 4.86E+03 | 1.61E+02 | 7.02E+03 | 0.00E+00 | 0.00E+00 | 7.27E+03 | 6.41E+03 | 7.10E+03 | 3.34E+03 | 2.98E+03 | 3.45E+03 | **0.00E+00** |
| | Mean | 0.00E+00 | 4.61E+03 | 7.88E+01 | 6.84E+03 | 0.00E+00 | 0.00E+00 | 6.93E+03 | 6.05E+03 | 6.75E+03 | 3.16E+03 | 2.73E+03 | 3.33E+03 | **0.00E+00** |
| | STD | 0.00e-F00 | 1.58E+02 | 2.83E+01 | 1.16E+02 | 0.00E+00 | 0.00E+00 | 1.65E+02 | 1.80E+02 | 1.24E+02 | 1.04E+02 | 1.16E+02 | 7.02E+01 | **0.00E+00** |
| F10 | Best | 8.88E-16 | 1.21E+01 | 1.23E-03 | 1.95E+01 | 8.88E-16 | 4.44E-15 | 2.03E+01 | 2.02E+01 | 1.91E+01 | 1.38E+01 | 1.01E+01 | 1.35E+01 | **4.44E-16** |
| | Worst | 8.88E-16 | 1.44E+01 | 3.38E-03 | 1.98E+01 | 7.99E-15 | 7.99E-15 | 2.05E+01 | 2.05E+01 | 2.01E+01 | 1.46E+01 | 1.11E+01 | 1.46E+01 | **4.44E-16** |
| | Mean | 8.88E-16 | 1.30E+01 | 1.90E-03 | 1.97E+01 | 4.44E-15 | 7.87E-15 | 2.04E+01 | 2.03E+01 | 1.95E+01 | 1.42E+01 | 1.05E+01 | 1.41E+01 | **4.44E-16** |
| | STD | 0.00e-F00 | 6.47E-01 | 4.40E-04 | 9.58E-02 | 2.46E-15 | 6.48E-16 | 1.49E-01 | 8.64E-02 | 1.45E-01 | 2.27E-01 | 2.32E-01 | 2.56E-01 | **0.00E+00** |
| F11 | Best | 0.00E+00 | 6.73E+02 | 1.07E-04 | 4.18E+03 | 0.00E+00 | 0.00E+00 | 9.83E+03 | 2.27E+03 | 4.02E+03 | 7.64E+02 | 8.17E+03 | 8.63E+01 | **0.00E+00** |
| | Worst | 0.00e-F00 | 1.28E+03 | 1.29E-01 | 5.16E+03 | 0.00E+00 | 0.00E+00 | 1.06E+04 | 3.30E+03 | 5.53E+03 | 9.47E+02 | 8.99E+03 | 1.11E+02 | **0.00E+00** |
| | Mean | 0.00e-F00 | 9.41E+02 | 3.35E-02 | 4.71E+03 | 0.00E+00 | 0.00E+00 | 1.02E+04 | 3.02E+03 | 5.03E+03 | 8.46E+02 | 8.618+03 | 9.618+01 | **0.00E+00** |
| | STD | 0.00E+00 | 1.74E+02 | 4.90E-02 | 2.14E+02 | 0.00E+00 | 0.00E+00 | 2.86E+02 | 2.62E+02 | 3.34E+02 | 5.56E+01 | 2.05E+02 | 5.73E+00 | **0.00E+00** |
| F12 | Best | 8.97E-09 | 8.55E+05 | 6.59E-01 | 4.04E+09 | 2.45E-02 | 5.89E-01 | 1.05E+10 | 3.84E+08 | 4.69E+09 | 2.27E+05 | 1.99E+02 | 1.39E+01 | **0.00E+00** |
| | Worst | 4.98E-04 | 1.47E+07 | 8.21E-01 | 9.90E+09 | 1.87E-01 | 7.19E-01 | 1.24E+10 | 7.24E+08 | 1.83E+10 | 3.69E+06 | 8.02E+04 | 2.42E+01 | **0.00E+00** |
| | Mean | 4.19E-05 | 4.05E+06 | 7.43E-01 | 7.44E+09 | 8.59E-02 | 6.52E-01 | 1.20E+10 | 4.95E+08 | 1.13E+10 | 1.39E+06 | 1.46E+04 | 1.92E+01 | **0.00E+00** |
| | STD | 9.45E-05 | 3.49E+06 | 4.36E-02 | 1.53E+09 | 4.24E-02 | 3.33E-02 | 5.00E+08 | 8.51E+07 | 4.24E+09 | 7.92E+05 | 1.91E+04 | 2.40E+00 | **0.00E+00** |
| F13 | Best | 3.38E-04 | 1.41E+07 | 4.88E+01 | 7.74E+09 | 9.45E+00 | 4.98E+01 | 1.94E+10 | 8.71E+08 | 7.35E+09 | 1.85E+07 | 1.69E+06 | 6.78E+03 | **5.00E+01** |
| | Worst | 1.99E-01 | 1.27E+08 | 5.40E+01 | 1.42E+10 | 2.58E+01 | 4.99E+01 | 2.42E+10 | 1.75E+09 | 3.43E+10 | 5.92E+07 | 7.18E+06 | 5.47E+04 | **5.00E+01** |
| | Mean | 2.52E-02 | 5.34E+07 | 5.11E+01 | 1.13E+10 | 1.80E+01 | 4.99E+01 | 2.21E+10 | 1.35E+09 | 1.47E+10 | 3.32E+07 | 3.89E+06 | 2.73E+04 | **5.00E+01** |
| | STD | 4.30E-02 | 2.83E+07 | 1.25E+00 | 1.63E+09 | 4.12E+00 | 7.42E-03 | 9.85E+08 | 2.47E+08 | 6.31E+09 | 8.49E+06 | 1.33E+08 | 1.13E+04 | **2.41E-03** |

Table 7 Results of benchmark functions (F1-F13), with 1000 dimensions

| No. | | AVOA | PSO | GWO | FFA | WOA | TLBO | MFO | BBO | DE | SSA | GSA | IPO | **SHMS** |
|---|---|---|---|---|---|---|---|---|---|---|---|---|---|---|
| F1 | Best | 7.50E-278 | 1.47E+05 | 1.41E-01 | 1.33E+06 | 9.06E-84 | 2.06E-87 | 2.62E+06 | 6.00E+05 | 1.49E+06 | 2.12E+05 | 1.22E+05 | 4.10E+04 | **0.00E+00** |
| | Worst | 3.06E-193 | 3.04E+05 | 4.79E-01 | 1.49E+06 | 5.41E-67 | 2.33E-84 | 2.83E+06 | 7.30E+05 | 1.81E+06 | 2.67E+05 | 1.48E+05 | 5.07E+04 | **0.00E+00** |
| | Mean | 1.05E-194 | 2.16E+05 | 2.42E-01 | 1.43E+06 | 1.80E-68 | 2.44E-85 | 2.73E+06 | 6.69E+05 | 1.60E+06 | 2.37E+05 | 1.31E+05 | 4.60E+04 | **0.00E+00** |
| | STD | 0.00E+00 | 4.33E+04 | 7.43E-02 | 4.49E+04 | 9.88E-68 | 4.40E-85 | 4.83E+04 | 2.99E+04 | 8.51E+04 | 1.30E+04 | 5.62E+03 | 2.32E+03 | **0.00E+00** |
| F2 | Best | 4.21E-173 | 1.00E+300 | 2.87E-01 | 1.00E+300 | 2.13E-56 | 1.00E+300 | 1.00E+300 | 1.00E+300 | 1.00E+300 | 1.11E+03 | 1.17E+263 | 4.39E+02 | **0.00E+00** |
| | Worst | 2.02E-112 | 1.00E+300 | 2.10E+00 | 1.00E+300 | 2.83E-47 | 1.00E+300 | 1.00E+300 | 1.00E+300 | 1.00E+300 | 1.25E+03 | 6.38E+289 | 4.88E+02 | **0.00E+00** |
| | Mean | 6.75E-114 | 1.00E+300 | 7.17E-01 | 1.00E+300 | 1.93E-48 | 1.00E+300 | 1.00E+300 | 1.00E+300 | 1.00E+300 | 1.19E+03 | 3.46E+288 | 4.60E+02 | **0.00E+00** |
| | STD | 3.69E-113 | 1.00E+300 | 3.96E-01 | 1.00E+300 | 5.68E-48 | 1.00E+300 | 1.00E+300 | 1.00E+300 | 1.00E+300 | 3.14E+01 | 1.00E+300 | 1.04E+01 | **0.00E+00** |
| F3 | Best | 2.27E-219 | 3.96E+06 | 1.15E+06 | 1.83E+07 | 6.83E+07 | 4.55E-06 | 1.31E+07 | 6.81E+06 | 3.62E+07 | 2.37E+06 | 2.99E+06 | 3.28E+05 | **0.00E+00** |
| | Worst | 5.26E-111 | 1.36E+07 | 2.65E+06 | 2.49E+07 | 2.26E+08 | 1.59E-01 | 2.66E+07 | 1.23E+07 | 5.94E+07 | 9.55E+06 | 1.37E+07 | 7.63E+05 | **0.00E+00** |
| | Mean | 1.76E-112 | 8.18E+06 | 1.67E+06 | 2.15E+07 | 1.32E+08 | 1.13E-02 | 1.85E+07 | 9.66E+06 | 4.77E+07 | 5.92E+06 | 6.53E+06 | 5.55E+05 | **0.00E+00** |
| | STD | 9.60E-112 | 2.44E+06 | 3.22E+05 | 1.82E+06 | 4.84E+07 | 3.16E-02 | 3.66E+06 | 1.60E+06 | 5.67E+06 | 1.92E+06 | 2.56E+06 | 9.61E+04 | **0.00E+00** |
| F4 | Best | 3.10E-143 | 3.61E+01 | 7.28E+01 | 9.91E+01 | 9.05E+00 | 2.61E-36 | 9.90E+01 | 8.27E+01 | 9.88E+01 | 4.00E+01 | 3.13E+01 | 2.24E+01 | **0.00E+00** |
| | Worst | 1.34E-100 | 5.58E+01 | 8.84E+01 | 9.95E+01 | 9.96E+01 | 1.50E-34 | 9.94E+01 | 8.99E+01 | 9.95E+01 | 5.26E+01 | 3.72E+01 | 2.47E+01 | **0.00E+00** |
| | Mean | 4.47E-102 | 4.23E+01 | 7.90E+01 | 9.92E+01 | 8.23E+01 | 2.97E-35 | 9.92E+01 | 8.60E+01 | 9.91E+01 | 4.48E+01 | 3.40E+01 | 2.35E+01 | **0.00E+00** |
| | STD | 2.44E-101 | 3.75E+00 | 3.21E+00 | 9.52E-02 | 2.01E+01 | 2.95E-35 | 1.42E-01 | 2.03E+00 | 1.40E-01 | 2.82E+00 | 1.28E+00 | 7.51E-01 | **0.00E+00** |
| F5 | Best | 1.44E-03 | 5.25E+07 | 1.00E+03 | 6.94E+09 | 9.92E+02 | 9.93E+02 | 1.19E+10 | 7.96E+08 | 1.42E+10 | 9.60E+07 | 2.09E+07 | 1.14E+07 | **2.89E+01** |
| | Worst | 5.25E+01 | 2.02E+08 | 1.12E+03 | 9.62E+09 | 9.95E+02 | 9.95E+02 | 1.31E+10 | 1.17E+09 | 1.53E+10 | 1.53E+08 | 2.85E+07 | 1.61E+07 | **2.90E+01** |
| | Mean | 5.80E+00 | 9.78E+07 | 1.02E+03 | 8.51E+09 | 9.93E+02 | 9.94E+02 | 1.24E+10 | 9.58E+08 | 1.48E+10 | 1.14E+08 | 2.47E+07 | 1.36E+07 | **2.90E+01** |
| | STD | 1.11E+01 | 3.42E+07 | 2.12E+01 | 6.35E+08 | 7.07E-01 | 1.31E-01 | 2.91E+08 | 8.91E+07 | 2.77E+08 | 1.21E+07 | 1.95E+06 | 1.09E+06 | **2.18E-02** |
| F6 | Best | 2.26E-03 | 1.41E+05 | 1.95E+02 | 1.28E+06 | 2.80E+01 | 2.03E+02 | 2.57E+06 | 6.07E+05 | 1.38E+06 | 2.15E+05 | 1.21E+05 | 4.69E+04 | **0.00E+00** |
| | Worst | 1.05E+00 | 2.91E+05 | 2.06E+02 | 1.47E+06 | 1.17E+02 | 2.18E+02 | 2.75E+06 | 7.21E+05 | 2.27E+06 | 2.54E+05 | 1.48E+05 | 9.84E+04 | **0.00E+00** |
| | Mean | 1.27E-01 | 1.99E+05 | 2.00E+02 | 1.41E+06 | 7.29E+01 | 2.13E+02 | 2.72E+06 | 6.66E+05 | 1.63E+06 | 2.37E+05 | 1.31E+05 | 6.24E+04 | **0.00E+00** |
| | STD | 2.06E-01 | 3.44E+04 | 2.79E+00 | 5.50E+04 | 1.91E+01 | 2.20E+00 | 4.95E+04 | 3.34E+04 | 1.61E+05 | 1.12E+04 | 5.65E+03 | 1.06E+04 | **0.00E+00** |
| F7 | Best | 1.41E-05 | 8.87E+02 | 9.98E-02 | 8.40E+04 | 1.30E-04 | 8.77E-04 | 1.90E+05 | 1.19E+04 | 9.92E+04 | 1.31E+03 | 5.43E+03 | 1.83E+04 | **2.60E-03** |
| | Worst | 9.11E-04 | 2.64E+03 | 2.10E-01 | 1.28E+05 | 1.87E-02 | 3.67E-03 | 2.09E+05 | 1.56E+04 | 2.49E+05 | 2.15E+03 | 8.26E+03 | 2.69E+04 | **2.84E-02** |
| | Mean | 2.85E-04 | 1.57E+03 | 1.48E-01 | 1.10E+05 | 3.54E-03 | 2.01E-03 | 1.97E+05 | 1.34E+04 | 2.05E+05 | 1.69E+03 | 6.43E+03 | 2.20E+04 | **1.02E-02** |
| | STD | 2.43E-04 | 4.93E+02 | 3.08E-02 | 1.27E+04 | 4.43E-03 | 6.18E-04 | 4.50E+03 | 1.15E+03 | 5.24E+04 | 1.74E+02 | 7.09E+02 | 2.00E+03 | **6.91E-03** |

| No. | | AVOA | PSO | GWO | FFA | WOA | TLBO | MFO | BBO | DE | SSA | GSA | IPO | **SHMS** |
|---|---|---|---|---|---|---|---|---|---|---|---|---|---|---|
| F8 | Best | -4.18E+05 | -3.43E+04 | -1.02E+05 | -6.79E+04 | -4.18E+05 | -9.18E+04 | -1.79E+05 | -8.37E+04 | -4.25E+04 | -1.01E+05 | -2.38E+04 | -7.88E+04 | **0.00E+00** |
| | Worst | -3.95E+05 | -1.98E+04 | -1.77E+04 | -2.46E+04 | -2.21E+05 | -2.75E+04 | -7.40E+04 | -7.22E+04 | -3.39E+04 | -7.11E+04 | -9.51E+03 | -1.60E+04 | **0.00E+00** |
| | Mean | -4.17E+05 | -2.58E+04 | -8.58E+04 | -3.72E+04 | -3.27E+05 | -5.87E+04 | -8.72E+04 | -7.80E+04 | -3.64E+04 | -8.72E+04 | -1.34E+04 | -5.47E+04 | **0.00E+00** |
| | STD | 4.69E+03 | 3.72E+03 | 1.88E+04 | 1.07E+04 | 6.28E+04 | 1.38E+04 | 6.85E+03 | 2.85E+03 | 2.00E+03 | 8.08E+03 | 2.78E+03 | 1.89E+04 | **0.00E+00** |
| F9 | Best | 0.00E+00 | 9.21E+03 | 1.11E+02 | 1.38E+04 | 0.00E+00 | 0.00E+00 | 1.50E+04 | 1.13E+04 | 1.38E+04 | 7.17E+03 | 6.23E+03 | 7.62E+03 | **0.00E+00** |
| | Worst | 0.00E+00 | 1.00E+04 | 2.90E+02 | 1.46E+04 | 0.00E+00 | 0.00E+00 | 1.58E+04 | 1.18E+04 | 1.43E+04 | 7.93E+03 | 7.02E+03 | 8.10E+03 | **0.00E+00** |
| | Mean | 0.00E+00 | 9.66E+03 | 1.90E+02 | 1.43E+04 | 0.00E+00 | 0.00E+00 | 1.52E+04 | 1.17E+04 | 1.41E+04 | 7.57E+03 | 6.69E+03 | 7.79E+03 | **0.00E+00** |
| | STD | 0.00E+00 | 1.87E+02 | 4.19E+01 | 1.64E+02 | 0.00E+00 | 0.00E+00 | 1.92E+02 | 2.42E+02 | 2.41E+02 | 2.07E+02 | 2.16E+02 | 1.13E+02 | **0.00E+00** |
| F10 | Best | 8.88E-16 | 1.24E+01 | 1.38E-02 | 1.95E+01 | 8.88E-16 | 7.99E-15 | 2.00E+01 | 1.96E+01 | 2.01E+01 | 1.42E+01 | 1.10E+01 | 1.41E+01 | **4.44E-16** |
| | Worst | 8.88E-16 | 1.55E+01 | 2.24E-02 | 2.03E+01 | 4.44E-15 | 1.30E+01 | 2.06E+01 | 2.03E+01 | 2.06E+01 | 1.49E+01 | 1.15E+01 | 1.48E+01 | **4.44E-16** |
| | Mean | 8.88E-16 | 1.37E+01 | 1.83E-02 | 2.00E+01 | 3.01E-15 | 4.29E-01 | 2.05E+01 | 2.00E+01 | 2.03E+01 | 1.46E+01 | 1.12E+01 | 1.45E+01 | **4.44E-16** |
| | STD | 0.00E+00 | 8.96E-01 | 2.17E-03 | 9.92E-02 | 1.77E-15 | 2.34E+00 | 1.91E-01 | 8.64E-02 | 1.10E-01 | 1.82E-01 | 1.43E-01 | 1.51E-01 | **0.00E+00** |
| F11 | Best | 0.00E+00 | 1.42E+03 | 8.47E-03 | 1.12E+04 | 0.00E+00 | 0.00E+00 | 2.40E+04 | 5.15E+03 | 1.24E+04 | 1.97E+03 | 1.98E+04 | 4.78E+02 | **0.00E+00** |
| | Worst | 0.00E+00 | 2.50E+03 | 2.50E-01 | 1.38E+04 | 0.00E+00 | 1.11E-16 | 2.51E+04 | 6.47E+03 | 1.70E+04 | 2.30E+03 | 2.13E+04 | 5.97E+02 | **0.00E+00** |
| | Mean | 0.00E+00 | 1.87E+03 | 5.11E-02 | 1.28E+04 | 0.00E+00 | 3.33E-17 | 2.46E+04 | 5.86E+03 | 1.45E+04 | 2.07E+03 | 2.05E+04 | 5.44E+02 | **0.00E+00** |
| | STD | 0.00E+00 | 2.73E+02 | 7.61E-02 | 5.92E+02 | 0.00E+00 | 5.17E-17 | 3.53E+02 | 3.55E+02 | 1.02E+03 | 8.45E+01 | 3.34E+02 | 2.69E+01 | **0.00E+00** |
| F12 | Best | 2.19E-07 | 1.28E+06 | 9.04E-01 | 2.38E+10 | 3.60E-02 | 8.20E-01 | 2.79E+10 | 7.51E+08 | 3.39E+10 | 5.05E+06 | 5.39E+04 | 2.68E+01 | **0.00E+00** |
| | Worst | 5.89E-04 | 3.48E+07 | 1.88E+00 | 3.85E+10 | 3.03E-01 | 8.94E-01 | 3.22E+10 | 1.07E+09 | 3.79E+10 | 1.79E+07 | 5.00E+05 | 1.90E+02 | **0.00E+00** |
| | Mean | 6.76E-05 | 1.09E+07 | 1.18E+00 | 3.17E+10 | 1.05E-01 | 8.54E-01 | 3.04E+10 | 8.87E+08 | 3.68E+10 | 1.11E+07 | 1.88E+05 | 5.47E+01 | **0.00E+00** |
| | STD | 1.44E-04 | 8.13E+06 | 2.76E-01 | 4.68E+09 | 6.06E-02 | 1.52E-02 | 1.01E+09 | 8.29E+07 | 1.09E+09 | 3.00E+06 | 1.19E+05 | 3.99E+01 | **0.00E+00** |
| F13 | Best | 1.93E-03 | 3.71E+07 | 1.06E+02 | 3.44E+10 | 1.91E+01 | 9.98E+01 | 5.15E+10 | 2.35E+09 | 6.49E+10 | 1.08E+08 | 9.50E+06 | 3.58E+05 | **1.00E+02** |
| | Worst | 2.89E-01 | 3.20E+08 | 1.24E+02 | 4.87E+10 | 6.70E+01 | 9.95E+01 | 5.80E+10 | 3.42E+09 | 6.92E+10 | 2.15E+08 | 2.66E+07 | 1.04E+06 | **1.00E+02** |
| | Mean | 5.76E-02 | 1.15E+08 | 1.18E+02 | 4.44E+10 | 3.65E+01 | 9.94E+01 | 5.55E+10 | 2.78E+09 | 6.70E+10 | 1.46E+08 | 1.60E+07 | 6.38E+05 | **1.00E+02** |
| | STD | 8.02E-02 | 6.11E+07 | 5.35E+00 | 3.17E+09 | 1.14E+01 | 1.17E-02 | 1.44E+09 | 2.71E+08 | 1.22E+09 | 2.20E+07 | 4.03E+06 | 1.92E+05 | **3.33E-03** |

Table 8 Results of benchmark functions (F14-F23)

| No. | | AVOA | PSO | GWO | FFA | WOA | TLBO | MFO | BBO | DE | SSA | GSA | IPO | **SHMS** |
|---|---|---|---|---|---|---|---|---|---|---|---|---|---|---|
| F14 | Best | 9.98E-01 | 9.98E-01 | 9.98E-01 | 9.98E-01 | 9.98E-01 | 9.98E-01 | 9.98E-01 | 9.98E-01 | 9.98E-01 | 9.98E-01 | 9.98E-01 | 9.98E-01 | **1.27E+01** |
| | Worst | 2.98E+00 | 1.60E+01 | 1.27E+01 | 7.83B+00 | 1.08E+01 | 9.98E-01 | 1.27E+01 | 7.87E+00 | 5.93E+00 | 4.95E+00 | 1.25E+01 | 6.90E+00 | **1.27E+01** |
| | Mean | 1.26E+00 | 5.95E+00 | 4.06E+00 | 1.91E+00 | 2.57E+00 | 9.98E-01 | 2.87E+00 | 3.27E+00 | 1.39E+00 | 1.39E+00 | 5.41E+00 | 2.61E+00 | **1.27E+01** |
| | STD | 5.79E-01 | 3.76E+00 | 4.18E+00 | 2.07E+00 | 2.54E+00 | 4.12E-17 | 2.55E+00 | 2.12E+00 | 1.26E+00 | 8.85E-01 | 3.35E+00 | 1.82E+00 | **0.00E+00** |
| F15 | Best | 3.08E-04 | 4.20E-04 | 3.07E-04 | 3.34B-04 | 3.08E-04 | 3.07E-04 | 7.12E-04 | 3.93E-04 | 3.08E-04 | 3.08E-04 | 1.17E-03 | 3.07E-04 | **4.67E-04** |
| | Worst | 7.57E-04 | 5.67E-02 | 2.04E-02 | 8.75E-04 | 2.13E-03 | 2.04E-02 | 2.01E-02 | 2.04E-02 | 2.01E-02 | 6.33E-02 | 1.29E-02 | 7.14E-04 | **7.65E-03** |
| | Mean | 4.65E-04 | 1.07E-02 | 4.39E-03 | 5.90E-04 | 6.55E-04 | 1.09E-03 | 1.93E-03 | 4.79E-03 | 1.14E-03 | 4.28E-03 | 4.35E-03 | 4.30E-04 | **2.75E-03** |
| | STD | 1.48E-04 | 1.80E-02 | 8.13E-03 | 1.19B-04 | 4.22E-04 | 3.65E-03 | 3.75E-03 | 7.92E-03 | 3.64E-03 | 1.22E-02 | 2.74E-03 | 1.36E-04 | **1.89E-03** |
| F16 | Best | -1.03E+00 | -1.03E+00 | -1.03E+00 | -1.03E+00 | -1.03E+00 | -1.03E+00 | -1.03E+00 | -1.03E+00 | -1.03E+00 | -1.03E+00 | -1.03E+00 | -1.03E+00 | **-1.03E+00** |
| | Worst | -1.03E+00 | -1.03E+00 | -1.03E+00 | -1.03E+00 | -1.03E+00 | -1.03E+00 | -1.03E+00 | -1.03E+00 | -1.03E+00 | -1.03E+00 | -1.03E+00 | -1.03E+00 | **-1.01E+00** |
| | Mean | -1.03E+00 | -1.03E+00 | -1.03E+00 | -1.03E+00 | -1.03E+00 | -1.03E+00 | -1.03E+00 | -1.03E+00 | -1.03E+00 | -1.03E+00 | -1.03E+00 | -1.03E+00 | **-1.03E+00** |
| | STD | 6.78E-16 | 6.59E-06 | 1.56E-08 | 6.78E-16 | 2.02E-09 | 6.78E-16 | 6.78E-16 | 4.43E-12 | 6.78E-16 | 4.09E-14 | 4.88E-16 | 5.61E-16 | **4.95E-03** |
| F17 | Best | 3.98E-01 | 3.98E-01 | 3.98E-01 | 3.98E-01 | 3.98E-01 | 3.98E-01 | 3.98E-01 | 3.98E-01 | 3.98E-01 | 3.98E-01 | 3.98E-01 | 3.98E-01 | **3.16E+00** |
| | Worst | 3.98E-01 | 8.86E-01 | 3.98E-01 | 3.98E-01 | 3.98E-01 | 3.98E-01 | 3.98E-01 | 8.92E-01 | 3.98E-01 | 3.98E-01 | 3.98E-01 | 3.98E-01 | **1.77E+01** |
| | Mean | 3.98E-01 | 4.62E-01 | 3.98E-01 | 3.98E-01 | 3.98E-01 | 3.98E-01 | 3.98E-01 | 4.77E-01 | 3.98E-01 | 3.98E-01 | 3.98E-01 | 3.98E-01 | **9.72E+00** |
| | STD | 1.69E-16 | 1.15E-01 | 1.47E-06 | 3.95E-12 | 6.06E-06 | 1.69E-16 | 1.69E-16 | 1.39E-01 | 1.69E-16 | 2.29E-05 | 1.69E-16 | 2.87E-13 | **3.31E+00** |
| F18 | Best | 3.00E+00 | 3.00E+00 | 3.00E+00 | 3.00E+00 | 3.00E+00 | 3.00E+00 | 3.00E+00 | 3.00E+00 | 3.00E+00 | 3.00E+00 | 3.00E+00 | 3.00E+00 | **3.22E+00** |
| | Worst | 3.00E+00 | 8.40E+01 | 3.00E+00 | 3.00E+00 | 3.00E+00 | 3.00E+00 | 3.00E+00 | 8.43E+01 | 3.00E+00 | 3.00E+00 | 3.00E+00 | 3.00E+00 | **3.21E+01** |
| | Mean | 3.00E+00 | 1.06E+01 | 3.00E+00 | 3.00E+00 | 3.00E+00 | 3.00E+00 | 3.00E+00 | 7.510+00 | 3.000+00 | 3.00E+00 | 3.00E+00 | 3.00E+00 | **8.90E+00** |
| | STD | 0.00E+00 | 1.72E+01 | 3.93E-05 | 0.00E+00 | 2.29E-04 | 0.00E+00 | 0.00E+00 | 1.60E+01 | 0.000+00 | 2.74E-12 | 4.20E-15 | 2.78E-15 | **7.24E+00** |
| F19 | Best | -3.86E+00 | -3.86E+00 | -3.86E+00 | -3.86E+00 | -3.86E+00 | -3.86E+00 | -3.86E+00 | -3.86E+00 | -3.86E+00 | -3.86E+00 | -3.86E+00 | -3.86E+00 | **-3.84E+00** |
| | Worst | -3.86E+00 | -3.32E+00 | -3.85E+00 | -3.86E+00 | -3.79E+00 | -3.86E+00 | -3.86E+00 | -3.86E+00 | -3.86E+00 | -3.86E+00 | -3.86E+00 | -3.86E+00 | **-3.13E+00** |
| | Mean | -3.86E+00 | -3.80E+00 | -3.86E+00 | -3.86E+00 | -3.85E+00 | -3.86E+00 | -3.86E+00 | -3.86E+00 | -3.86E+00 | -3.86E+00 | -3.86E+00 | -3.86E+00 | **-3.59E+00** |
| | STD | 9.16E-10 | 1.05E-01 | 2.45E-03 | 3.48E-08 | 1.60E-02 | 2.63E-15 | 1.44E-03 | 3.42E-15 | 2.71E-15 | 5.26E-10 | 2.31E-15 | 2.25E-15 | **2.05E-01** |
| F20 | Best | -3.32E+00 | -3.29E+00 | -3.32E+00 | -3.32E+00 | -3.32E+00 | -3.32E+00 | -3.32E+00 | -3.32E+00 | -3.32E+00 | -3.32E+00 | -3.32E+00 | -3.32E+00 | **-2.55E+00** |
| | Worst | -3.20E+00 | -1.30E+00 | -3.09E+00 | -3.20E+00 | -3.07E+00 | -3.20E+00 | -3.14E+00 | -3.20E+00 | -3.20E+00 | -3.15E+00 | -3.32E+00 | -3.20E+00 | **-1.08E+00** |
| | Mean | -3.31E+00 | -2.77E+00 | -3.27E+00 | -3.30E+00 | -3.19E+00 | -3.29E+00 | -3.22E+00 | -3.28E+00 | -3.29E+00 | -3.23E+00 | -3.32E+00 | -3.31E+00 | **-1.82E+00** |
| | STD | 3.02E-02 | 4.31E-01 | 7.55E-02 | 4.43E-02 | 9.04E-02 | 5.34E-02 | 5.20E-02 | 5.83E-02 | 5.11E-02 | 6.33E-02 | 1.61E-15 | 3.03E-02 | **3.92E-01** |

| No. | | AVOA | PSO | GWO | FFA | WOA | TLBO | MFO | BBO | DE | SSA | GSA | IPO | **SHMS** |
|---|---|---|---|---|---|---|---|---|---|---|---|---|---|---|
| F21 | Best | -1.02E+01 | -1.01E+01 | -1.02E+01 | -1.02E+01 | -1.02E+01 | -1.02E+01 | -1.02E+01 | -1.02E+01 | -1.02E+01 | -1.02E+01 | -1.02E+01 | -1.02E+01 | **-1.69E+00** |
| | Worst | -1.02E+05 | -6.87E-01 | -2.68E+00 | -4.81E+00 | -2.63E+00 | -2.68E+00 | -2.63E+00 | -2.63E+00 | -2.68E+00 | -2.63E+00 | -2.63E+00 | -2.63E+00 | **-9.28E-01** |
| | Mean | -1.02E+01 | -3.78E+00 | -8.15E+00 | -9.02E+00 | -7.66E+00 | -9.26E+00 | -5.56E+00 | -5.07E+00 | -9.40E+00 | -6.30E+00 | -7.03E+00 | -8.07E+00 | **-1.25E+00** |
| | STD | 1.08E-10 | 2.86E+00 | 2.95E+00 | 1.96E+00 | 2.90E+00 | 2.02E+00 | 3.41E+00 | 3.24E+00 | 1.99E+00 | 3.53E+00 | 3.68E+00 | 3.29E+00 | **2.42E-01** |
| F22 | Best | -1.04E+01 | -1.04E+01 | -1.04E+01 | -1.04E+01 | -1.04E+01 | -1.04E+01 | -1.04E+01 | -1.04E+01 | -1.04E+01 | -1.04E+01 | -1.04E+01 | -1.04E+01 | **-3.49E+00** |
| | Worst | -1.04E+01 | -1.60E+00 | -1.04E+01 | -3.68E+00 | -2.76E+00 | -3.61E+00 | -2.75E+00 | -2.75E+00 | -2.75E+00 | -2.75E+00 | -2.77E+00 | -2.75E+00 | **-9.40E-01** |
| | Mean | -1.04E+01 | -5.04E+00 | -1.04E+01 | -9.68E+00 | -7.79E+00 | -8.70E+00 | -9.05E+00 | -5.96E+00 | -9.85E+00 | -8.89E+00 | -9.79E+00 | -9.89E+00 | **-1.34E+00** |
| | STD | 9.14E-11 | 2.92E+00 | 1.32E-03 | 2.05E+00 | 3.09E+00 | 2.67E+00 | 2.78E+00 | 3.47E+00 | 1.85E+00 | 2.850+00 | 1.90E+00 | 1.94E+00 | **5.06E-01** |
| F23 | Best | -1.05E+01 | -1.05E+01 | -1.05E+01 | -1.05E+01 | -1.05E+01 | -1.05E+01 | -1.05E+01 | -1.05E+01 | -1.05E+01 | -1.05E+01 | -1.05E+01 | -1.05E+01 | **-1.94E+00** |
| | Worst | -1.05E+01 | -1.63E+00 | -2.42E+00 | -2.87E+00 | -7.89E-01 | -3.84E+00 | -2.43E+00 | -1.86E+00 | -2.87E+00 | -2.43E+00 | -2.43E+00 | -2.42E+00 | **-9.80E-01** |
| | Mean | -1.05E+01 | -4.89E+00 | -1.03E+01 | -9.81E+00 | -6.81E+00 | -9.87E+00 | -8.68E+00 | -5.28E+00 | -1.03E+01 | -8.53E+00 | -9.49E+00 | -7.81E+00 | **-1.35E+00** |
| | STD | 4.34E-11 | 3.20E+00 | 1.48E+00 | 2.14E+00 | 3.34E+00 | 2.04E+00 | 3.16E+00 | 3.55E+00 | 1.40E+00 | 3.42E+00 | 2.72E+00 | 3.49E+00 | **2.43E-01** |

Table 9: Comparison of average running time results *(sec.)* over 30 runs for larger-scale problems with 1000 variables

| No. | | AVOA | PSO | GWO | FFA | WOA | TLBO | MFO | BBO | DE | SSA | GSA | IPO | **SHMS** |
|---|---|---|---|---|---|---|---|---|---|---|---|---|---|---|
| F1 | Best | 9.17E-01 | 1.45E+00 | 2.70E+00 | 2.77E+00 | 2.51E+00 | 1.40E+00 | 1.85E+00 | 1.40E+01 | 1.65E+00 | 1.56E+00 | 1.61E+01 | 6.13E+00 | **1.38E-01** |
| | Worst | 1.08E+00 | 2.40E+00 | 3.16E+00 | 3.24E+00 | 2.90E+00 | 2.39E+00 | 2.12E+00 | 1.50E+01 | 2.08E+00 | 1.76E+00 | 1.88E+01 | 9.62E+00 | **1.69E-01** |
| | Mean | 1.02E+00 | 1.64E+00 | 2.75E+00 | 2.85E+00 | 2.74E+00 | 1.63E+00 | 1.93E+00 | 1.42E+01 | 1.73E+00 | 1.63E+00 | 1.64E+01 | 6.73E+00 | **1.45E-01** |
| | STD | 7.82E-02 | 2.19E-01 | 9.87E-02 | 1.31E-01 | 1.08E-01 | 2.14E-01 | 6.81E-02 | 1.48E-01 | 1.12E-01 | 4.87E-02 | 6.90E-01 | 6.83E-01 | **7.26E-03** |
| F2 | Best | 9.92E-01 | 1.51E+00 | 2.79E+00 | 2.99E+00 | 2.81E+00 | 1.50E+00 | 1.90E+00 | 1.42E+01 | 1.69E+00 | 1.58E+00 | 6.56E+01 | 6.00E+00 | **8.43E-02** |
| | Worst | 1.13E+00 | 1.86E+00 | 3.32E+00 | 4.33E+00 | 3.18E+00 | 1.88E+00 | 2.24E+00 | 1.87E+01 | 1.99E+00 | 1.78E+00 | 9.04E+01 | 7.19E+00 | **9.66E-02** |
| | Mean | 1.01E+00 | 1.72E+00 | 2.89E+00 | 3.27E+00 | 2.96E+00 | 1.65E+00 | 2.00E+00 | 1.46E+01 | 1.89E+00 | 1.62E+00 | 7.30E+01 | 6.51E+00 | **9.03E-02** |
| | STD | 4.36E-02 | 1.01E-01 | 1.04E-01 | 3.16E-01 | 9.40E-02 | 8.26E-02 | 8.58E-02 | 8.72E-01 | 8.18E-02 | 4.77E-02 | 7.88E+00 | 2.70E-01 | **3.88E-03** |
| F3 | Best | 2.40E+01 | 3.02E+01 | 3.05E+01 | 5.43E+01 | 3.06E+01 | 5.53E+01 | 2.84E+01 | 5.85E+01 | 2.96E+01 | 2.82E+01 | 3.94E+01 | 2.19E+01 | **1.26E+00** |
| | Worst | 2.54E+01 | 3.66E+01 | 3.71E+01 | 5.95E+01 | 3.50E+01 | 6.07E+01 | 2.98E+01 | 5.93E+01 | 3.35E+01 | 3.00E+01 | 4.19E+01 | 3.90E+01 | **1.68E+00** |
| | Mean | 2.41E+01 | 3.10E+01 | 3.10E+01 | 5.47E+01 | 3.13E+01 | 5.60E+01 | 2.89E+01 | 5.86E+01 | 3.01E+01 | 2.84E+01 | 3.97E+01 | 2.95E+01 | **1.50E+00** |
| | STD | 3.34E-01 | 1.49E+00 | 1.45E+00 | 1.11E+00 | 9.16E-01 | 1.03E+00 | 2.16E-01 | 3.10E-01 | 8.12E-01 | 3.33E-01 | 6.33E-01 | 5.66E+00 | **1.15E-01** |
| F4 | Best | 8.99E-01 | 1.42E+00 | 2.74E+00 | 2.63E+00 | 2.82E+00 | 1.35E+00 | 1.82E+00 | 1.40E+01 | 1.71E+00 | 1.54E+00 | 1.61E+01 | 6.03E+00 | **7.72E-01** |
| | Worst | 1.07E+00 | 2.44E+00 | 3.25E+00 | 3.27E+00 | 3.82E+00 | 2.13E+00 | 2.00E+00 | 1.64E+01 | 2.23E+00 | 1.71E+00 | 1.98E+01 | 6.64E+00 | **1.12E+00** |

| No. | | AVOA | PSO | GWO | FFA | WOA | TLBO | MFO | BBO | DE | SSA | GSA | IPO | **SHMS** |
|---|---|---|---|---|---|---|---|---|---|---|---|---|---|---|
| | Mean | 9.56E-01 | 1.69E+00 | 2.86E+00 | 2.84E+00 | 3.04E+00 | 1.47E+00 | 1.89E+00 | 1.43E+01 | 1.89E+00 | 1.59E+00 | 1.64E+01 | 6.45E+00 | **9.81E-01** |
| | STD | 3.88E-02 | 2.00E-01 | 1.00E-01 | 1.47E-01 | 2.05E-01 | 1.77E-01 | 5.74E-02 | 5.73E-01 | 1.45E-01 | 4.92E-02 | 9.44E-01 | 2.02E-01 | **8.36E-02** |
| | Best | 9.97E-01 | 1.76E+00 | 2.80E+00 | 2.84E+00 | 2.88E+00 | 1.54E+00 | 1.91E+00 | 1.40E+01 | 1.64E+00 | 1.66E+00 | 1.62E+01 | 6.32E+00 | **7.76E-01** |
| F5 | Worst | 1.19E+00 | 2.09E+00 | 4.97E+00 | 3.33E+00 | 3.71E+00 | 2.12E+00 | 2.09E+00 | 1.56E+01 | 2.32E+00 | 1.87E+00 | 2.14E+01 | 6.87E+00 | **9.10E-01** |
| | Mean | 1.03E+00 | 1.91E+00 | 3.10E+00 | 3.03E+00 | 3.21E+00 | 1.75E+00 | 1.97E+00 | 1.42E+01 | 1.91E+00 | 1.74E+00 | 1.66E+01 | 6.72E+00 | **8.19E-01** |
| | STD | 4.27E-02 | 1.06E-01 | 4.31E-01 | 9.62E-02 | 1.90E-01 | 1.68E-01 | 6.04E-02 | 3.38E-01 | 1.57E-01 | 5.85E-02 | 1.34E+00 | 9.59E-02 | **3.17E-02** |
| | Best | 9.38E-01 | 1.45E+00 | 2.73E+00 | 2.74E+00 | 2.94E+00 | 1.34E+00 | 1.83E+00 | 1.41E+01 | 1.69E+00 | 1.62E+00 | 1.61E+01 | 5.89E+00 | **9.65E-02** |
| F6 | Worst | 1.11E+00 | 2.02E+00 | 3.46E+00 | 3.17E+00 | 3.77E+00 | 1.96E+00 | 2.00E+00 | 1.51E+01 | 2.21E+00 | 1.90E+00 | 1.93E+01 | 6.58E+00 | **1.16E-01** |
| | Mean | 9.74E-01 | 1.76E+00 | 2.88E+00 | 2.88E+00 | 3.14E+00 | 1.52E+00 | 1.90E+00 | 1.43E+01 | 1.91E+00 | 1.70E+00 | 1.67E+01 | 6.36E+00 | **1.03E-01** |
| | STD | 3.80E-02 | 1.29E-01 | 1.82E-01 | 9.69E-02 | 1.58E-01 | 1.60E-01 | 5.20E-02 | 1.33E-01 | 1.34E-01 | 6.84E-02 | 1.03E+00 | 1.45E-01 | **3.68E-03** |
| | Best | 2.94E+00 | 3.97E+00 | 5.20E+00 | 7.25E+00 | 5.08E+00 | 6.17E+00 | 4.21E+00 | 1.51E+01 | 4.06E+00 | 3.97E+00 | 1.82E+01 | 7.24E+00 | **9.05E-02** |
| F7 | Worst | 3.41E+00 | 4.54E+00 | 5.52E+00 | 7.75E+00 | 6.75E+00 | 6.56E+00 | 4.51E+00 | 1.65E+01 | 4.37E+00 | 4.27E+00 | 2.60E+01 | 1.03E+01 | **1.79E-01** |
| | Mean | 3.05E+00 | 4.07E+00 | 5.27E+00 | 7.37E+00 | 5.37E+00 | 6.24E+00 | 4.25E+00 | 1.55E+01 | 4.12E+00 | 4.02E+00 | 2.01E+01 | 7.86E+00 | **1.08E-01** |
| | STD | 1.11E-01 | 1.88E-01 | 8.60E-02 | 1.25E-01 | 3.59E-01 | 8.83E-02 | 6.89E-02 | 3.20E-01 | 8.51E-02 | 6.51E-02 | 3.27E+00 | 6.37E-01 | **1.79E-02** |
| | Best | 1.26E+00 | 2.45E+00 | 3.17E+00 | 3.89E+00 | 3.33E+00 | 2.51E+00 | 2.28E+00 | 1.50E+01 | 2.60E+00 | 2.19E+00 | 2.38E+01 | 6.12E+00 | **9.65E-02** |
| F8 | Worst | 1.77E+00 | 3.51E+00 | 3.74E+00 | 4.71E+00 | 3.92E+00 | 3.19E+00 | 2.77E+00 | 1.59E+01 | 3.64E+00 | 2.54E+00 | 2.88E+01 | 6.94E+00 | **1.16E-01** |
| | Mean | 1.35E+00 | 2.71E+00 | 3.23E+00 | 4.25E+00 | 3.59E+00 | 2.78E+00 | 2.41E+00 | 1.52E+01 | 2.82E+00 | 2.32E+00 | 2.45E+01 | 6.62E+00 | **1.03E-01** |
| | STD | 1.19E-01 | 1.84E-01 | 1.08E-01 | 1.68E-01 | 1.38E-01 | 1.72E-01 | 1.13E-01 | 2.54E-01 | 2.25E-01 | 7.37E-02 | 1.34E+00 | 1.57E-01 | **3.68E-03** |
| | Best | 1.02E+00 | 2.26E+00 | 2.93E+00 | 3.72E+00 | 3.05E+00 | 1.88E+00 | 2.18E+00 | 1.45E+01 | 2.53E+00 | 2.03E+00 | 1.63E+01 | 6.25E+00 | **2.73E-01** |
| F9 | Worst | 1.26E+00 | 3.00E+00 | 3.23E+00 | 4.51E+00 | 3.68E+00 | 2.58E+00 | 2.67E+00 | 1.55E+01 | 2.89E+00 | 2.31E+00 | 2.38E+01 | 9.95E+00 | **3.48E-01** |
| | Mean | 1.07E+00 | 2.48E+00 | 3.00E+00 | 4.02E+00 | 3.29E+00 | 2.17E+00 | 2.28E+00 | 1.50E+01 | 2.62E+00 | 2.13E+00 | 1.91E+01 | 6.87E+00 | **3.00E-01** |
| | STD | 4.70E-02 | 1.56E-01 | 7.73E-02 | 2.07E-01 | 1.41E-01 | 1.78E-01 | 9.77E-02 | 1.69E-01 | 6.92E-02 | 7.15E-02 | 3.54E+00 | 6.97E-01 | **2.03E-02** |
| | Best | 1.07E+00 | 2.10E+00 | 2.98E+00 | 3.75E+00 | 3.05E+00 | 1.96E+00 | 2.26E+00 | 1.46E+01 | 2.58E+00 | 2.08E+00 | 1.64E+01 | 4.77E+00 | **1.22E-01** |
| F10 | Worst | 1.54E+00 | 2.77E+00 | 3.27E+00 | 4.86E+00 | 3.69E+00 | 2.61E+00 | 2.52E+00 | 1.56E+01 | 3.12E+00 | 2.35E+00 | 1.95E+01 | 5.42E+00 | **1.51E-01** |
| | Mean | 1.16E+00 | 2.34E+00 | 3.01E+00 | 4.00E+00 | 3.31E+00 | 2.24E+00 | 2.32E+00 | 1.50E+01 | 2.74E+00 | 2.18E+00 | 1.66E+01 | 4.95E+00 | **1.33E-01** |
| | STD | 9.40E-02 | 1.90E-01 | 8.05E-02 | 2.43E-01 | 1.32E-01 | 1.88E-01 | 5.85E-02 | 2.08E-01 | 1.34E-01 | 7.08E-02 | 7.95E-01 | 1.09E-01 | **6.94E-03** |
| | Best | 1.23E+00 | 2.27E+00 | 3.10E+00 | 4.16E+00 | 3.27E+00 | 2.24E+00 | 2.38E+00 | 1.46E+01 | 2.81E+00 | 2.25E+00 | 1.66E+01 | 6.52E+00 | **8.38E-02** |
| F11 | Worst | 1.56E+00 | 3.50E+00 | 3.70E+00 | 5.20E+00 | 4.40E+00 | 2.91E+00 | 2.69E+00 | 1.55E+01 | 3.26E+00 | 2.53E+00 | 1.73E+01 | 1.03E+01 | **1.05E-01** |
| | Mean | 1.29E+00 | 2.60E+00 | 3.21E+00 | 4.42E+00 | 3.51E+00 | 2.48E+00 | 2.49E+00 | 1.50E+01 | 2.92E+00 | 2.35E+00 | 1.67E+01 | 7.19E+00 | **9.31E-02** |
| | STD | 8.44E-02 | 2.20E-01 | 1.45E-01 | 2.19E-01 | 2.00E-01 | 1.76E-01 | 7.97E-02 | 2.29E-01 | 1.14E-01 | 6.92E-02 | 1.69E-01 | 6.23E-01 | **4.97E-03** |

| No. | | AVOA | PSO | GWO | FFA | WOA | TLBO | MFO | BBO | DE | SSA | GSA | IPO | **SHMS** |
|---|---|---|---|---|---|---|---|---|---|---|---|---|---|---|
| F12 | Best | 3.62E+00 | 5.27E+00 | 6.16E+00 | 9.41E+00 | 6.24E+00 | 8.40E+00 | 5.24E+00 | 1.53E+01 | 5.66E+00 | 5.00E+00 | 1.89E+01 | 8.98E+00 | **8.38E-02** |
| | Worst | 4.41E+00 | 6.35E+00 | 6.68E+00 | 1.02E+01 | 6.66E+00 | 9.02E+00 | 5.63E+00 | 2.12E+01 | 6.17E+00 | 5.47E+00 | 2.23E+01 | 9.70E+00 | **1.05E-01** |
| | Mean | 3.76E+00 | 5.44E+00 | 6.29E+00 | 9.80E+00 | 6.39E+00 | 8.54E+00 | 5.31E+00 | 1.58E+01 | 5.79E+00 | 5.09E+00 | 1.94E+01 | 9.45E+00 | **9.31E-02** |
| | STD | 1.83E-01 | 2.55E-01 | 1.35E-01 | 1.49E-01 | 9.46E-02 | 1.24E-01 | 8.54E-02 | 1.01E+00 | 1.20E-01 | 9.87E-02 | 9.67E-01 | 1.19E-01 | **4.97E-03** |
| F13 | Best | 3.62E+00 | 5.33E+00 | 6.09E+00 | 9.33E+00 | 6.18E+00 | 7.96E+00 | 5.18E+00 | 1.54E+01 | 5.51E+00 | 5.09E+00 | 1.90E+01 | 9.90E+00 | **2.65E-01** |
| | Worst | 4.42E+00 | 6.37E+00 | 6.73E+00 | 1.00E+01 | 6.83E+00 | 8.49E+00 | 5.67E+00 | 1.65E+01 | 6.40E+00 | 5.50E+00 | 1.93E+01 | 1.07E+01 | **2.99E-01** |
| | Mean | 3.77E+00 | 5.48E+00 | 6.20E+00 | 9.68E+00 | 6.41E+00 | 8.15E+00 | 5.27E+00 | 1.58E+01 | 5.82E+00 | 5.19E+00 | 1.91E+01 | 1.01E+01 | **2.80E-01** |
| | STD | 1.96E-01 | 2.54E-01 | 1.24E-01 | 1.66E-01 | 1.27E-01 | 1.22E-01 | 1.04E-01 | 2.07E-01 | 2.07E-01 | 9.25E-02 | 7.58E-02 | 1.33E-01 | **1.30E-02** |

**Table 10 Pairwise Wilcoxon signed rank test for benchmark functions (F1-13), with 30 dimensions**

| Other Algorithms vs SNAIL | p-value | T+ | T- | Winner |
|---|---|---|---|---|
| AVOA vs SNAIL | 9.66E-01 | 32 | 34 | **SHMS** |
| PSO vs SNAIL | 1.05E-02 | 10 | 81 | **SHMS** |
| GWO vs SNAIL | 8.02E-01 | 41.5 | 49.5 | **SHMS** |
| FFA vs SNAIL | 9.42E-02 | 21 | 70 | **SHMS** |
| WOA vs SNAIL | 6.22E-01 | 32 | 46 | **SHMS** |
| TLBO vs SNAIL | 8.93E-01 | 43 | 48 | **SHMS** |
| MFO vs SNAIL | 1.34E-02 | 11 | 80 | **SHMS** |
| BBO vs SNAIL | 9.24E-02 | 21 | 70 | **SHMS** |
| DE vs SNAIL | 8.03E-02 | 20 | 71 | **SHMS** |
| SSA vs SNAIL | 2.15E-02 | 13 | 78 | **SHMS** |
| GSA vs SNAIL | 2.15E-02 | 13 | 78 | **SHMS** |
| IPO vs SNAIL | 1.27E-01 | 23 | 68 | **SHMS** |

**Table 11 Pairwise Wilcoxon signed rank test for benchmark functions (F1-13), with 100 dimensions**

| Other Algorithms vs SNAIL | p-value | T+ | T- | Winner |
|---|---|---|---|---|
| AVOA vs SNAIL | 1.34E-02 | 11 | 80 | **SHMS** |
| PSO vs SNAIL | 4.60E-03 | 7 | 84 | **SHMS** |
| GWO vs SNAIL | 2.15E-02 | 13 | 78 | **SHMS** |
| FFA vs SNAIL | 8.10E-03 | 9 | 82 | **SHMS** |
| WOA vs SNAIL | 1.02E-01 | 14 | 52 | **SHMS** |
| TLBO vs SNAIL | 1.75E-01 | 17 | 49 | **SHMS** |
| MFO vs SNAIL | 4.60E-03 | 7 | 84 | **SHMS** |
| BBO vs SNAIL | 1.71E-02 | 12 | 79 | **SHMS** |
| DE vs SNAIL | 8.10E-03 | 9 | 82 | **SHMS** |
| SSA vs SNAIL | 1.34E-02 | 11 | 80 | **SHMS** |
| GSA vs SNAIL | 6.10E-03 | 8 | 83 | **SHMS** |
| IPO vs SNAIL | 1.71E-02 | 12 | 79 | **SHMS** |

**Table 12 Pairwise Wilcoxon signed rank test for benchmark functions (F1-13), with 500 dimensions**

| Other Algorithms vs SNAIL | p-value | T+ | T- | Winner |
|---|---|---|---|---|
| AVOA vs SNAIL | 7.65E-01 | 37 | 29 | AVOA |
| PSO vs SNAIL | 4.60E-03 | 7 | 84 | **SHMS** |
| GWO vs SNAIL | 1.71E-02 | 12 | 79 | **SHMS** |
| FFA vs SNAIL | 3.40E-03 | 6 | 85 | **SHMS** |
| WOA vs SNAIL | 2.78E-01 | 20 | 46 | **SHMS** |
| TLBO vs SNAIL | 4.65E-01 | 24 | 42 | **SHMS** |
| MFO vs SNAIL | 3.40E-03 | 6 | 85 | **SHMS** |
| BBO vs SNAIL | 8.10E-03 | 9 | 82 | **SHMS** |
| DE vs SNAIL | 4.60E-03 | 7 | 84 | **SHMS** |
| SSA vs SNAIL | 4.60E-03 | 7 | 84 | **SHMS** |
| GSA vs SNAIL | 2.40E-03 | 5 | 86 | **SHMS** |
| IPO vs SNAIL | 1.34E-02 | 11 | 80 | **SHMS** |

**Table 13 Pairwise Wilcoxon signed rank test for benchmark functions (F1-13), with 1000 dimensions**

| Other Algorithms vs SNAIL | p-value | T+ | T- | Winner |
|---|---|---|---|---|
| AVOA vs SNAIL | 7.65E-01 | 37 | 29 | AVOA |
| PSO vs SNAIL | 3.40E-03 | 6 | 85 | **SHMS** |
| GWO vs SNAIL | 1.71E-02 | 12 | 79 | **SHMS** |
| FFA vs SNAIL | 2.40E-03 | 5 | 86 | **SHMS** |
| WOA vs SNAIL | 2.78E-01 | 20 | 46 | **SHMS** |
| TLBO vs SNAIL | 2.04E-01 | 22 | 56 | **SHMS** |
| MFO vs SNAIL | 2.40E-03 | 5 | 86 | **SHMS** |
| BBO vs SNAIL | 3.40E-03 | 6 | 86 | **SHMS** |
| DE vs SNAIL | 2.40E-03 | 5 | 86 | **SHMS** |
| SSA vs SNAIL | 4.60E-03 | 7 | 84 | **SHMS** |
| GSA vs SNAIL | 2.40E-03 | 5 | 86 | **SHMS** |
| IPO vs SNAIL | 8.10E-03 | 9 | 82 | **SHMS** |

**Table 14 Pairwise Wilcoxon signed rank test for benchmark functions (F14-23)**

| Other Algorithms vs SNAIL | p-value | T+ | T- | Winner |
|---|---|---|---|---|
| AVOA vs SNAIL | 3.90E-03 | 45 | 0 | AVOA |
| PSO vs SNAIL | 3.91E-02 | 40 | 5 | PSO |
| GWO vs SNAIL | 7.80E-03 | 44 | 1 | GWO |
| FFA vs SNAIL | 3.90E-03 | 45 | 0 | FFA |
| WOA vs SNAIL | 3.90E-03 | 45 | 0 | WOA |
| TLBO vs SNAIL | 3.90E-03 | 45 | 0 | TLBO |
| MFO vs SNAIL | 3.90E-03 | 45 | 0 | MFO |
| BBO vs SNAIL | 7.80E-03 | 44 | 1 | BBO |
| DE vs SNAIL | 3.90E-03 | 45 | 0 | DE |
| SSA vs SNAIL | 7.80E-03 | 44 | 1 | SSA |
| GSA vs SNAIL | 7.80E-03 | 44 | 1 | GSA |
| IPO vs SNAIL | 3.90E-03 | 45 | 0 | IPO |

Table 15 Friedmann test for benchmark functions (F1-F13), with 30 dimensions

| Algorithm | AVOA | PSO | GWO | FFA | WOA | TLBO | MFO | BBO | DE | SSA | GSA | IPO | **SHMS** |
|---|---|---|---|---|---|---|---|---|---|---|---|---|---|
| Mean Values | 2.5 | 11.6923 | 5.2308 | 7.3077 | 5.8462 | 3.6923 | 11.5385 | 7.7692 | 7.5385 | 8.6154 | 8.7692 | 7.2308 | **3.2692** |
| Ranking | 1 | 13 | 4 | 7 | 5 | 3 | 12 | 9 | 8 | 10 | 11 | 6 | **2** |

Table 16 Friedmann test for benchmark functions (F1-F13), with 100 dimensions

| Algorithm | AVOA | PSO | GWO | FFA | WOA | TLBO | MFO | BBO | DE | SSA | GSA | IPO | **SHMS** |
|---|---|---|---|---|---|---|---|---|---|---|---|---|---|
| Mean Values | 1.8462 | 10.6923 | 4.8462 | 10.4615 | 4.3846 | 3.5385 | 11.7692 | 6.4615 | 10.7692 | 8.2308 | 8.6923 | 6.8462 | **2.4615** |
| Ranking | 1 | 11 | 5 | 10 | 4 | 3 | 13 | 6 | 12 | 8 | 9 | 7 | **2** |

Table 17 Friedmann test for benchmark functions (F1-F13), with 500 dimensions

| Algorithm | AVOA | PSO | GWO | FFA | WOA | TLBO | MFO | BBO | DE | SSA | GSA | IPO | **SHMS** |
|---|---|---|---|---|---|---|---|---|---|---|---|---|---|
| Mean Values | 1.7692 | 8.7692 | 5.2308 | 11.0385 | 4.1538 | 3.6923 | 11.8462 | 7.9231 | 11.3462 | 7.6154 | 8.2308 | 6.6923 | **2.6923** |
| Ranking | 1 | 10 | 5 | 11 | 4 | 3 | 13 | 8 | 12 | 7 | 9 | 6 | **2** |

Table 18 Friedmann test for benchmark functions (F1-F13), with 1000 dimensions

| Algorithm | AVOA | PSO | GWO | FFA | WOA | TLBO | MFO | BBO | DE | SSA | GSA | IPO | **SHMS** |
|---|---|---|---|---|---|---|---|---|---|---|---|---|---|
| Mean Values | 1.7308 | 8.0385 | 5 | 10.9615 | 4.0385 | 4.3077 | 11.4231 | 9.5385 | 11.8077 | 7.5769 | 7.5385 | 6.3846 | **2.6538** |
| Ranking | 1 | 9 | 5 | 11 | 3 | 4 | 12 | 10 | 13 | 8 | 7 | 6 | **2** |

Table 19 Friedmann test for benchmark functions (F14-F23)

| Algorithm | AVOA | PSO | GWO | FFA | WOA | TLBO | MFO | BBO | DE | SSA | GSA | IPO | **SHMS** |
|---|---|---|---|---|---|---|---|---|---|---|---|---|---|
| Mean Values | 13.35 | 11.6 | 6.15 | 5.05 | 7.7 | 5.1 | 7.25 | 9.65 | 4.7 | 7 | 6.45 | 5.2 | **11.8** |
| Ranking | 13 | 11 | 5 | 4 | 9 | 2 | 8 | 10 | 1 | 7 | 6 | 3 | **12** |

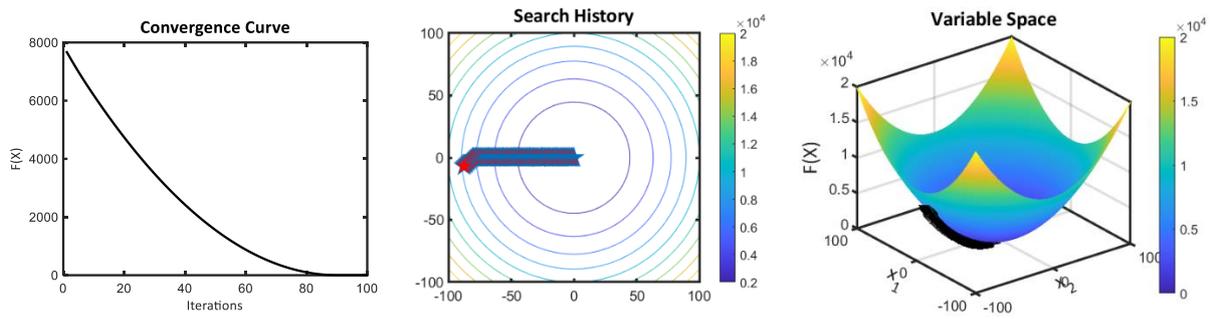

**Fig. 3(a): Sphere function**

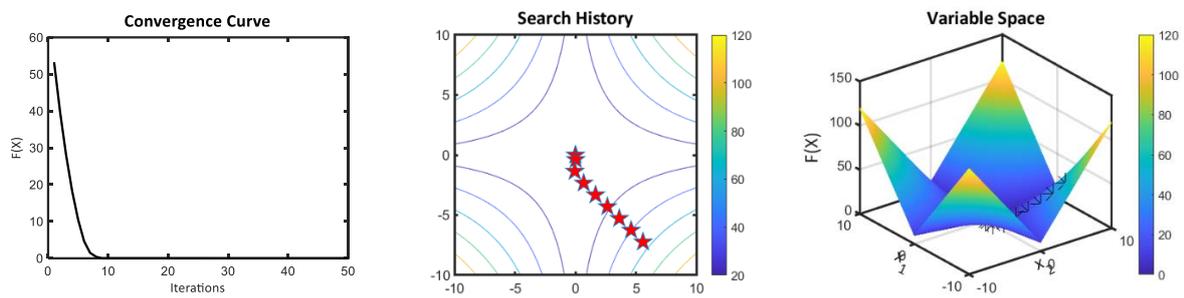

**Fig. 3(b): Schwefel 2.22 function**

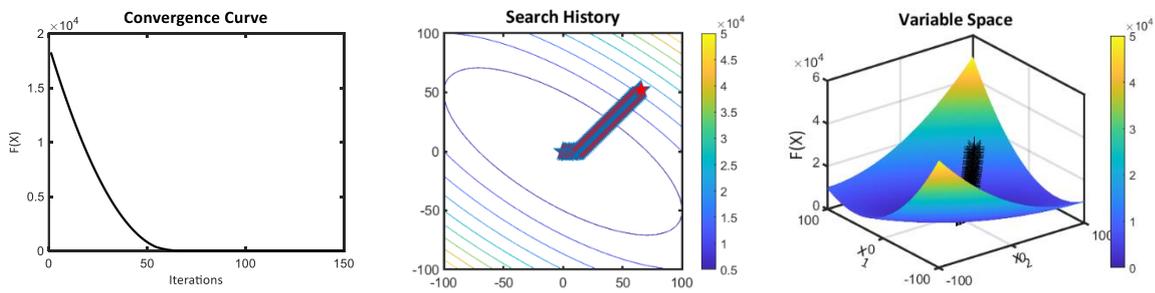

**Fig. 3(c): Schwefel 1.2 function**

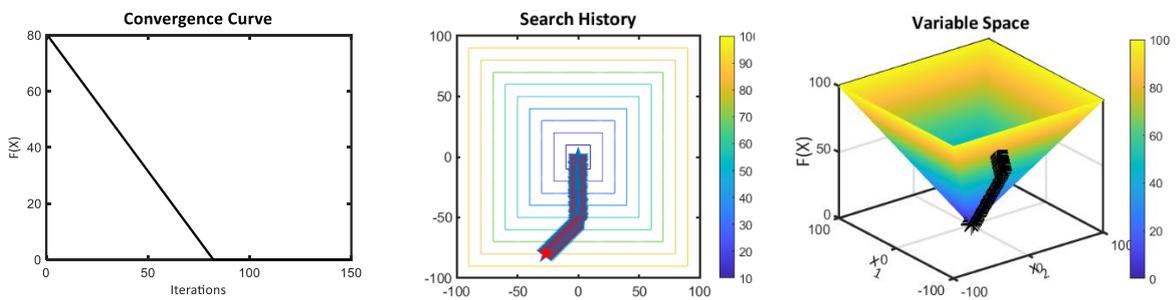

**Fig. 3(d): Schwefel 2.21 function**

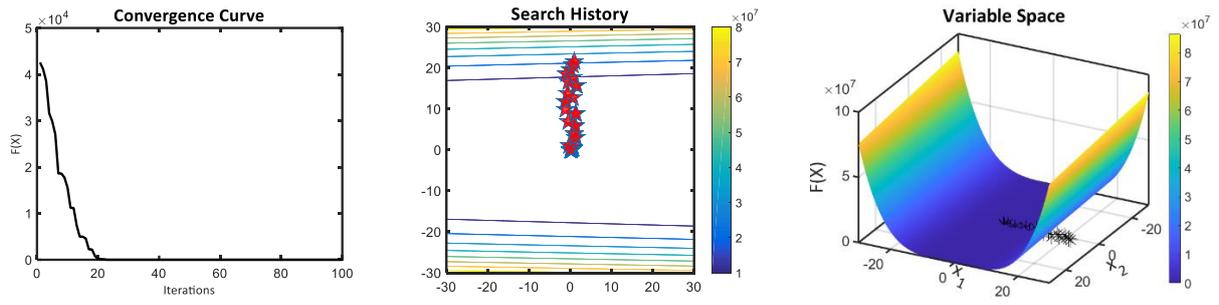

Fig. 3(e): Rosenbrock function

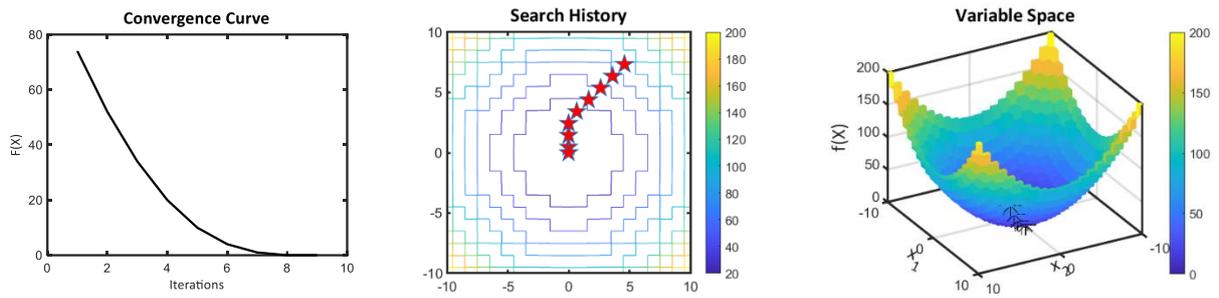

Fig. 3(f): Step function

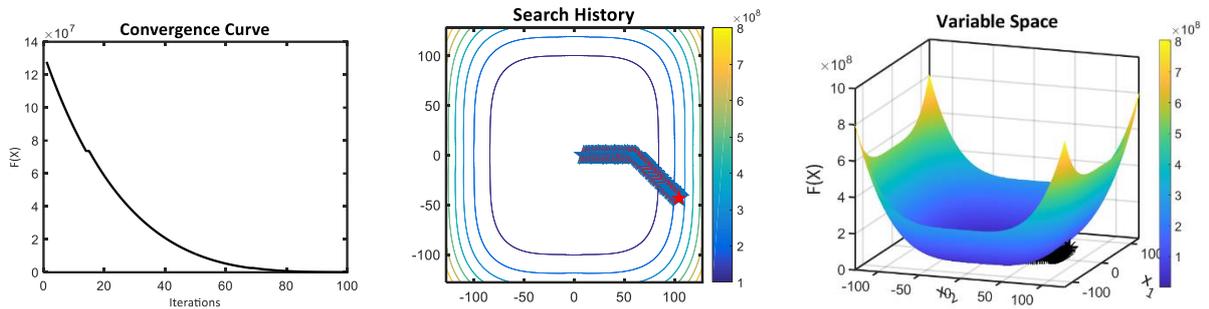

Fig. 3(e): Quartic function

Figure 3: The convergence curve, contour plot and 3D mesh plot of SHMS results solving unimodal benchmark test problems

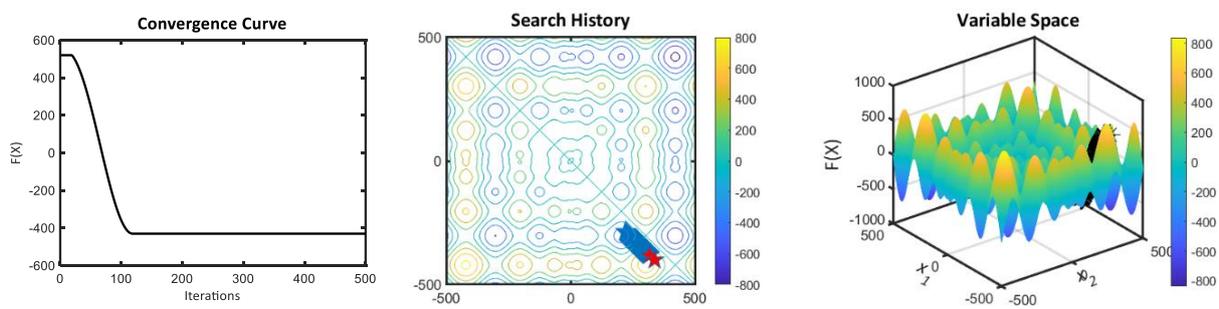

Fig. 4(a): Schwefel function

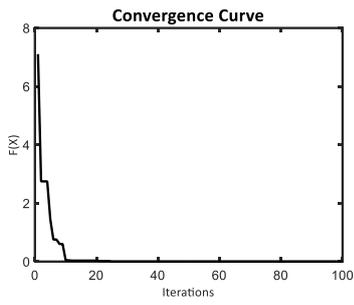 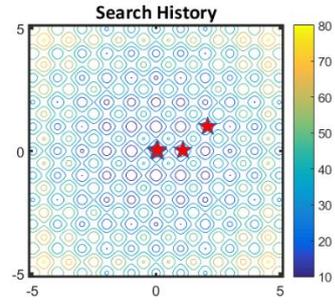 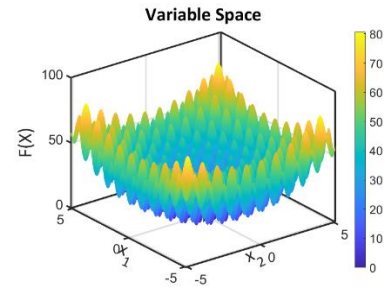

**Fig. 4 (b): Rastrigin function**

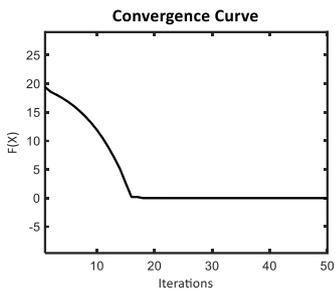 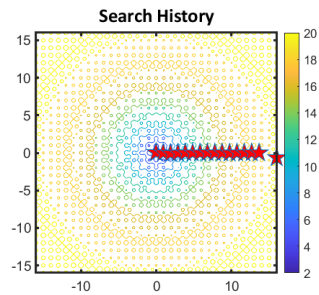 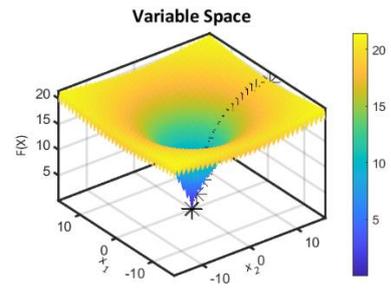

**Fig. 4 (c): Ackley function**

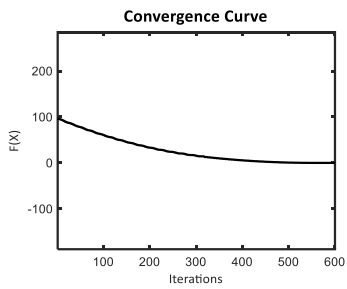 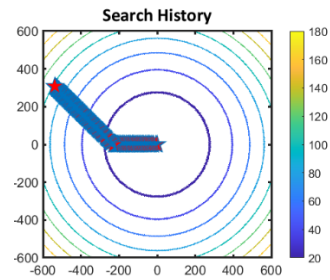 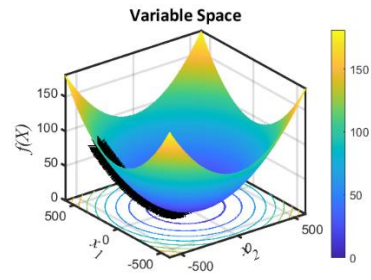

**Fig. 4 (d): Griewank function**

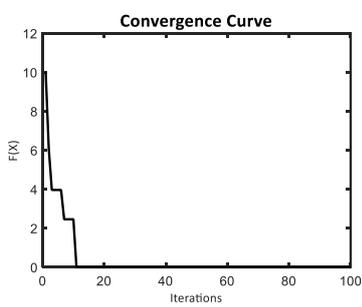 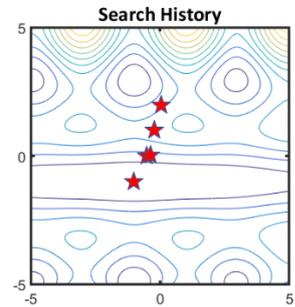 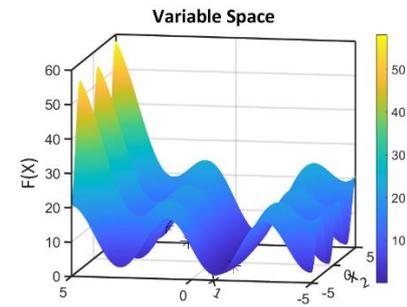

**Fig. 4 (e): Penalized function**

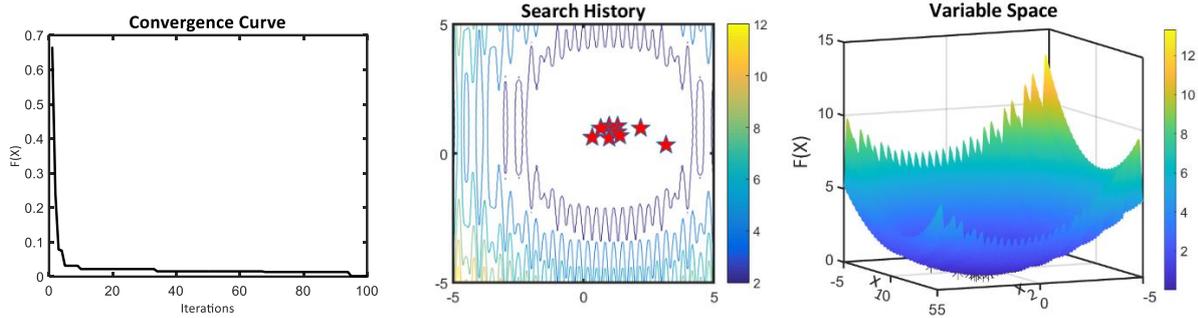

Fig. 4 (f): Penalized 2 function

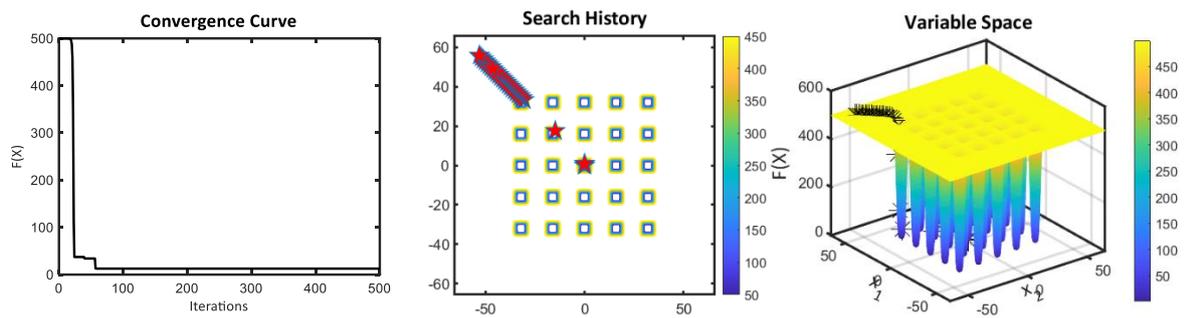

Fig. 4 (g): Foxholes function

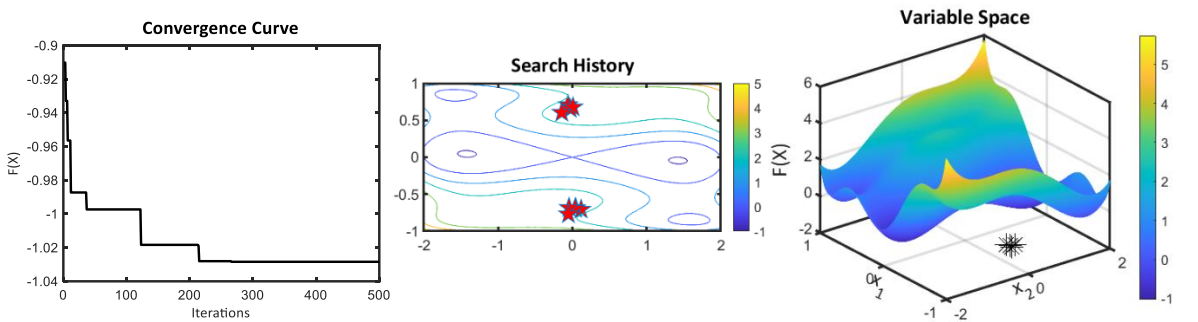

Fig. 4 (h): Six Hump Camel function

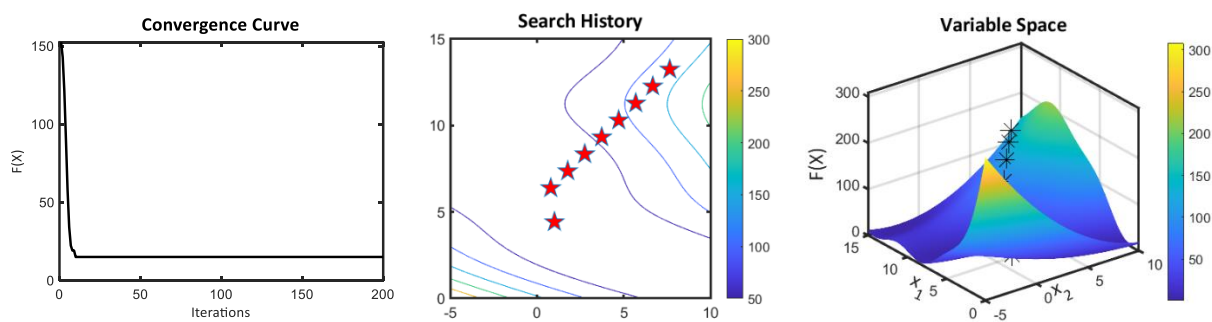

Fig. 4 (i): Branin function

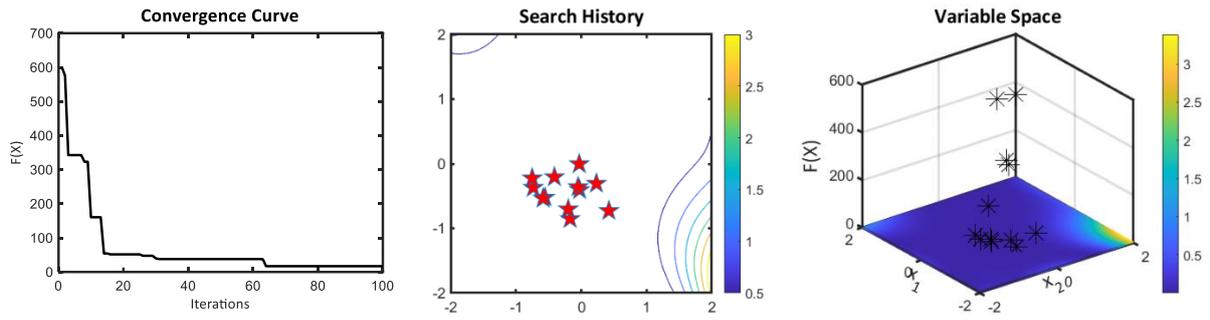

**Fig. 4 (i): Goldstein Price function**

**Fig 4: The convergence curve, contour plot and 3D mesh plot of SHMS results solving multimodal benchmark test problems**

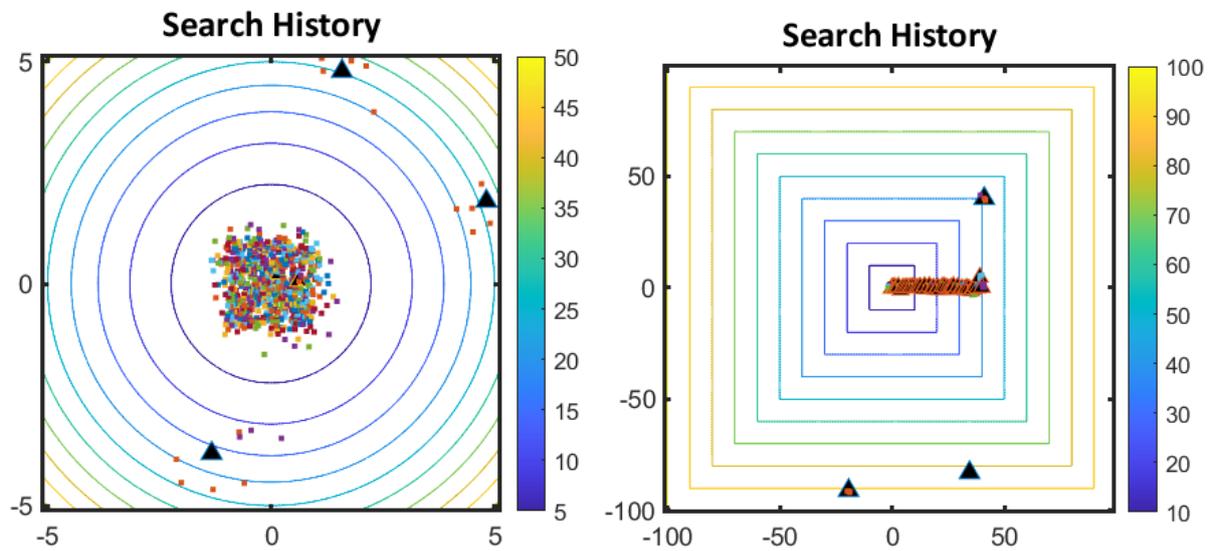

(a) Sphere function  (b) Schwefel 2.21 function

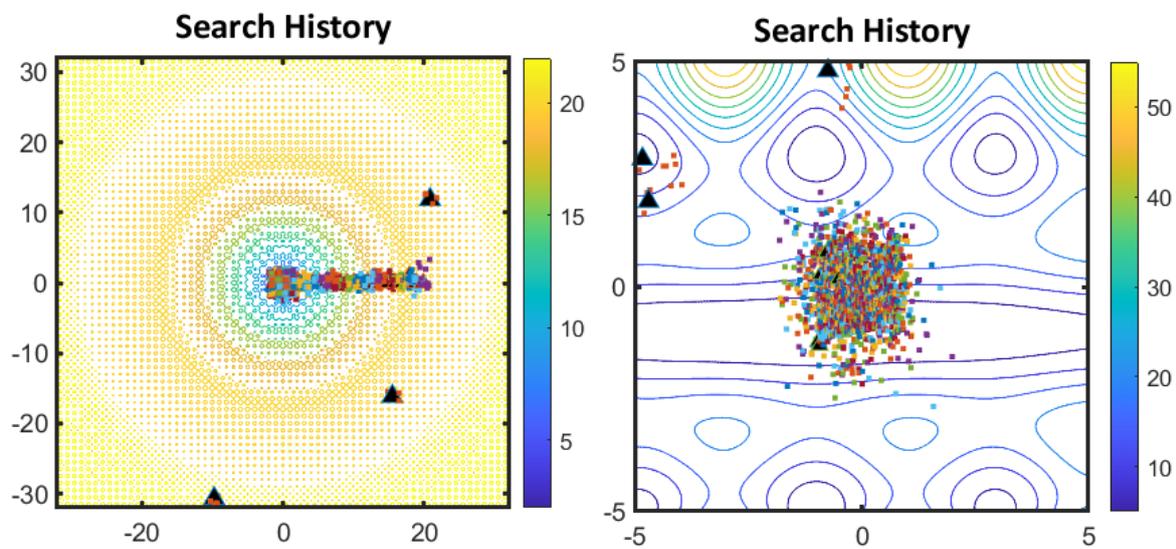

(c) Ackley function  (d) Penalized function

Fig 5: The performance of SHMS algorithm representing the homes and associated snails

## 5. Solution to Shell and Tube heat exchanger design and economic optimization problem

The shell and tube heat exchanger problem is most widely used to test an applicability of the metaheuristic algorithms. This problem has three cases as follows:

**Case 1:** 4.34 (MW) heat duty, methanol-brackish water heat exchanger.

**Case 2:** 1.44 (MW) heat duty, kerosene-crude oil heat exchanger.

**Case 3:** 0.46 (MW) heat duty, distilled water-raw water heat exchanger.

The problem has been previously solved using several metaheuristic algorithms such as GA (Caputo et al. 2008), PSO (Patel and Rao 2010), ABC (Sahin et al.2011), BBO (Hadidi and Nazari 2013), Intelligent Tuned Harmony Search Algorithm (ITHS), Improved Intelligent Tuned Harmony Search Algorithm (I-ITHS) (Turgut et al. 2014), CI (Dhavle et al. 2016), FFA (Mohanty et al. 2016), Adaptive Range Genetic Algorithm (ARGA) (Iyer et al., 2019), TLBO (Rao and Saroj, 2017) and SAMPE-JAYA (Rao and Saroj, 2017) algorithm. The design of the STHE on the mentioned costs was considered: the capital investment ($C_{inv}$), annual operating cost ($C_{annual}$), energy cost ($C_E$) and total discounted operating cost ($C_{total\_disc}$) (Caputo et al. 2008). The objective is to minimize the total cost ($C_{total}$). The mathematical formulation of the design and economic optimization of the shell and tube heat exchanger problem is adopted from ARGA (Iyer et al., 2019). The work comprehensively addresses all three cases of this problem.

The SHMS algorithm is successfully investigated for solving three cases of shell and tube heat exchanger problem. The performance from SHMS algorithm is compared with well-known nature inspired optimization algorithms such as GA, PSO, ABC, BBO, ITHS, I-ITHS, FFA, CI, TLBO, SAMPE-Jaya and ARGA. The result comparison of Case 1, Case 2 and Case 3 of shell and tube heat exchanger problem are presented in Table 20, Table 21 and Table 22, respectively. The comparison tables illustrate the outstanding performance of the SHMS algorithm in Case 1 and Case 2 when compared with the other metaheuristic algorithms presented in Table 20 and Table 21. However, in Case 3, the SHMS algorithm fell short of achieving a better solution when compared to the TLBO, SAMPLE-JAYA, I-ITHS, ITHS, BBO, and ABC algorithms. The cost comparison associated with these cases are presented in Figure 6, Figure 8 and Figure 10, respectively. The SHMS algorithm is run for 30 times. The statistical results such as best, mean and worst function value, standard deviation, function evaluations and CPU time are presented in Table 23. The closeness to the best reported solution (%) is presented in Table 24. Snails are known for their ability to sense humidity gradients and move towards areas with higher humidity, which is vital for their survival. Similarly, in optimization algorithms, the idea is to mimic or draw inspiration from such natural processes to solve complex problems. The convergence plots for all the three cases obtained from the SHMS algorithm are presented in Figure 7, Figure 9 and Figure 11.

**For Case 1:** Solving shell and tube heat exchanger problem case 1, the best, mean and worst function values ($C_{total}$) obtained from 30 trails are 41718.6558, 41725.3892 and 41728.6558, respectively with standard deviation 4.0847, average function evaluations 20510 and CPU time 9.62 $sec$. The best solution obtained using SHMS algorithm is 0.4649% better as compared to previously best reported solution using ARGA (Iyer et al., 2019) (refer Table 20).

**For Case 2:** The best mean worst function values ($C_{total}$) using SHMS algorithm for Case 2 are 19084.3059, 19088.3476 and 19097.2054, respectively with standard deviation 3.1663, function evaluations 17235 and CPU time 8.10 $sec$. The best solution obtained using SHMS algorithm is 0.5952% better as compared to previously best reported solution using ARGA (Iyer et al., 2019) (refer Table 21).

**For Case 3:** The best mean worst function values ($C_{total}$) using SHMS algorithm for Case 3 are 20744.3639 20746.1280 and 20749.8314, respectively with standard deviation 1.45653.1663, function evaluations 44721 and CPU time 20.13 $sec$. The best solution obtained using SHMS algorithm is 0.2775% better as compared to previously best reported solution using ARGA (Iyer et al., 2019) (refer Table 22).

SHMS algorithm has indeed demonstrated superior performance compared to other metaheuristic algorithms for solving the shell and tube heat exchanger design and economic optimization problem. It is due to its ability to efficiently navigate through the solution space, guided by the principles of cue following observed in snails. This makes it particularly effective in solving problems that involve multivariable and complex nature.

**Table 20: Comparison of SHMS results Design and Economic Optimization of Shell and Tube Heat Exchanger Problem Case 1**

| Parameters | Original Study | GA | PSO | ABC | BBO | ITHS | I-ITHS | CI | FFA | TLBO | SAMPE-Jaya | ARGA | SHMS |
|---|---|---|---|---|---|---|---|---|---|---|---|---|---|
| $D_s(m)$ | 0.894 | 0.83 | 0.81 | 1.3905 | 0.801 | 0.762 | 0.7635 | 0.7800 | 0.858 | 0.858 | 0.76860 | 0.6651 | **0.6447** |
| $L(m)$ | 4.83 | 3.379 | 3.115 | 3.963 | 2.04 | 2.0791 | 2.0391 | 1.9367 | 2.416 | 2.416 | 1.47660 | 1.2636 | **1.1121** |
| $b(m)$ | 0.356 | 0.5 | 0.424 | 0.4669 | 0.5 | 0.4988 | 0.4955 | 0.500 | 0.402 | 0.402 | 0.4999 | 0.4903 | **0.4166** |
| $d_o(m)$ | 0.02 | 0.016 | 0.015 | 0.0104 | 0.01 | 0.0101 | 0.01 | 0.010 | 0.01575 | 0.01575 | 0.01 | 0.01 | **0.01** |
| $P_t(m)$ | 0.025 | 0.02 | 0.0187 | - | 0.0125 | 0.1264 | 0.0125 | 0.0125 | 0.01968 | 0.01968 | 0.0125 | 0.0125 | **0.0125** |
| $C_1(m)$ | 0.005 | 0.004 | 0.0037 | - | 0.0025 | 0.0253 | 0.0025 | 0.0025 | - | | | 0.0025 | **0.0025** |
| $n_t$ | 2 | 2 | 2 | 2 | 2 | 2 | 2 | 2 | 2 | | | 2 | **2** |
| $N_t$ | 918 | 1567 | 1658 | 1528 | 3587 | 3454 | 3558 | 3734.1233 | 1692 | 1692 | 3614 | 2625.873 | **2451.7768** |
| $v_t(m/s)$ | 0.75 | 0.69 | 0.67 | 0.36 | 0.77 | 0.782 | 0.7744 | 0.7381 | 0.656 | 0.656 | 0.7624 | 1.0492 | **1.1237** |
| $Re_t$ | 14925 | 10936 | 10503 | - | 7642.49 | 7842.52 | 7701.29 | 7342.7474 | 10286 | 10286 | 7586.57 | 10440.12 | **11181.4577** |
| $Pr_t$ | 5.7 | 5.7 | 5.7 | - | 5.7 | 5.7 | 5.7 | 5.6949 | 5.7 | 5.7 | 5.7 | 5.694915 | **5.6949** |
| $h_t(W/m^2K)$ | 3812 | 3762 | 3721 | 3818 | 4314 | 4415.918 | 4388.79 | 4584.7085 | 6228 | 6228 | 3777.88 | 6196.002 | **6545.5440** |
| $f_t$ | 0.028 | 0.031 | 0.0311 | - | 0.034 | 0.0354 | 0.03555 | 0.0343 | 0.03119 | 0.03119 | 0.03401 | 0.0310 | **0.0305** |
| $\Delta P_t(Pa)$ | 6251 | 4298 | 4171 | 3043 | 6156 | 6998.7 | 6887.63 | 5862.7287 | 4246 | 4246 | 5078.37 | 9756.238 | **10349.6306** |
| $a_s(m^2)$ | 0.032 | 0.083 | 0.0687 | - | 0.0801 | 0.07602 | 0.07567 | 0.0780 | - | | | 0.0652 | **0.0537** |
| $D_e(m)$ | 0.014 | 0.011 | 0.0107 | - | 0.007 | 0.00719 | 0.00711 | 0.0071 | 0.0105 | 0.0105 | 0.00711 | 0.0071 | **0.0071** |
| $v_s(m/s)$ | 0.58 | 0.44 | 0.53 | 0.118 | 0.46 | 0.48755 | 0.48979 | 0.4752 | 0.54 | 0.54 | 0.48221 | 0.5683 | **0.6899** |
| $Re_s$ | 18381 | 11075 | 12678 | - | 7254 | 7736.89 | 7684.054 | 7451.3906 | 12625 | 12625 | 7571.34 | 8912.325 | **10819.8036** |
| $Pr_s$ | 5.1 | 5.1 | 5.1 | - | 5.1 | 5.08215 | 5.08215 | 5.0821 | 5.1 | 5.1 | 5.1 | 5.0821 | **5.0821** |
| $h_s(W/m^2K)$ | 1573 | 1740 | 1950.8 | 3396 | 2197 | 2213.89 | 2230.913 | 2195.9461 | 1991 | 1991 | 2084.05 | 2422.804 | **2695.5285** |
| $f_s$ | 0.33 | 0.357 | 0.349 | - | 0.379 | 0.3759 | 0.37621 | 0.3780 | 0.349 | 0.349 | 0.377126 | 0.3680 | **0.3575** |
| $\Delta P_s(Pa)$ | 35789 | 13267 | 20551 | 8390 | 13799 | 14794.94 | 14953.91 | 13608.4472 | 18788 | 18788 | 10488.39 | 10746.31 | **15447.1357** |
| $U (W/m^2K)$ | 615 | 660 | 713.9 | 832 | 755 | 760.594 | 761.578 | 764.5084 | 876.4 | 876.4 | 719.05 | 1031.472 | **1090.5668** |
| $S (m^2)$ | 278.6 | 262.8 | 243.2 | - | 229.95 | 228.32 | 228.03 | 227.1607 | 202.3 | 202.3 | 167.56 | 168.2758 | **159.1575** |
| $C_{inv}(€)$ | 51507 | 49259 | 46453 | 44559 | 44536 | 44301.66 | 44259.01 | 44132.5190 | 39336 | 39336 | 37519.89 | 35498.87 | **34139.5254** |
| $C_{annual}(€/year)$ | 21111 | 947 | 1038.7 | 1014.5 | 984 | 964.164 | 962.4858 | 955.9112 | 1040 | 1040 | 731.71 | 1043.96 | **1233.4685** |
| $C_{total\_disc}(€)$ | 12973 | 5818 | 6778.2 | 6233.8 | 6046 | 5924.343 | 5914.058 | 5873.6607 | 6446 | 6446 | 4496.08 | 6414.68 | **7579.1304** |
| $C_{total}(€)$ | 64480 | 55077 | 53231 | 50793 | 50582 | 50226 | 50173 | 50006.1797 | 45782 | 45,782 | 42015.98 | 41913.54 | **41718.6558** |

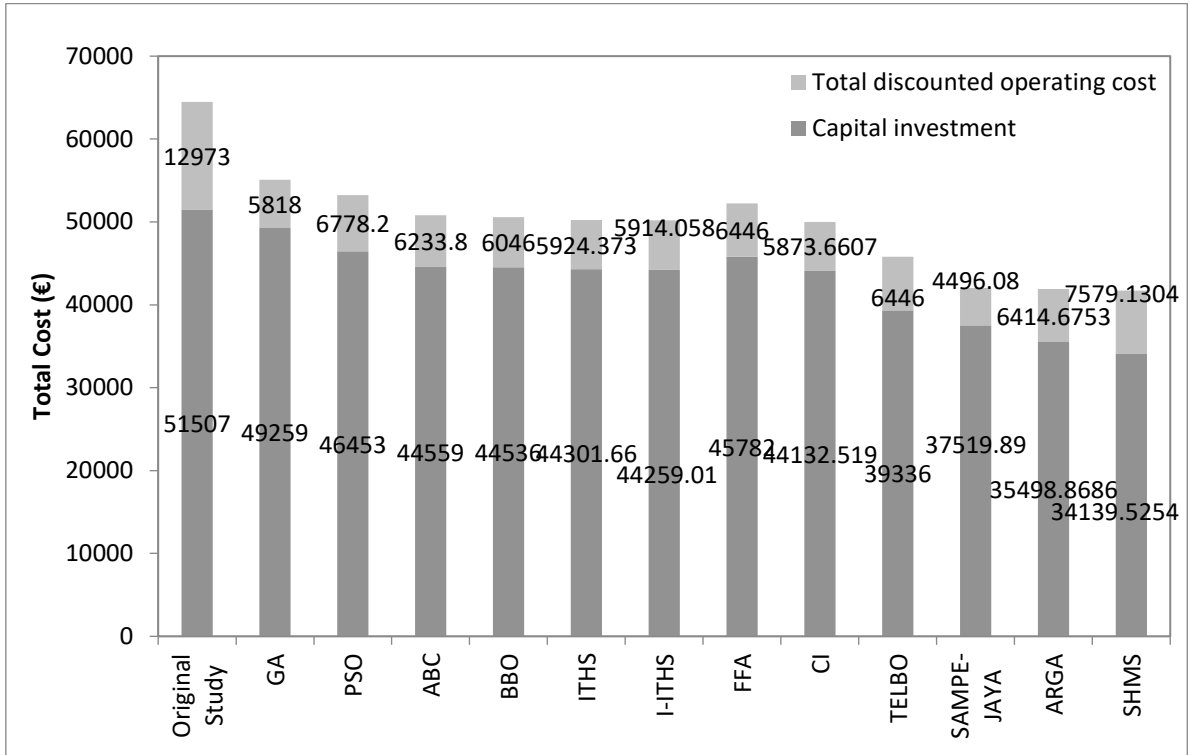

**Fig. 6: Total Cost Comparison for Case 1**

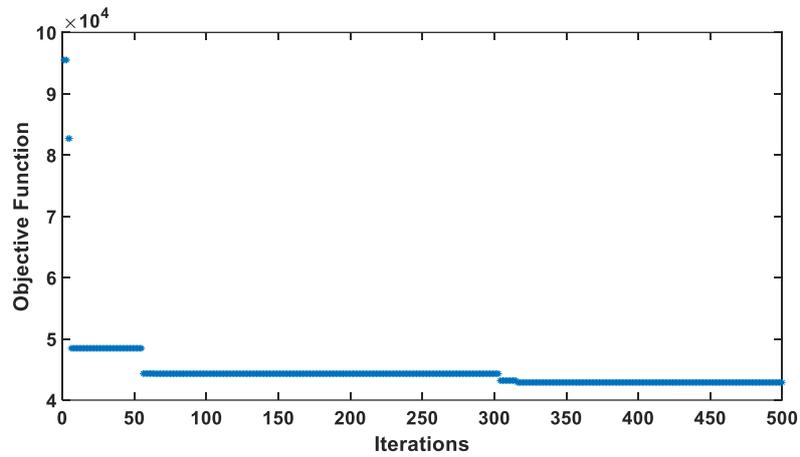

**Fig. 7: Convergence plot for Design and Economic Optimization of Shell and Tube Heat Exchanger Problem Case 1**

Table 21: Comparison of SHMS results Design and Economic Optimization of Shell and Tube Heat Exchanger Problem Case 2

| Parameters | Original Study | GA | PSO | ABC | BBO | ITHS | I-ITHS | CI | FFA | ARGA | **SHMS** |
|---|---|---|---|---|---|---|---|---|---|---|---|
| $D_s(m)$ | 0.539 | 0.63 | 0.59 | 0.3293 | 0.74 | 0.32079 | 0.31619 | 0.4580 | 0.7276 | 0.400009 | **0.4** |
| $L(m)$ | 4.88 | 2.153 | 1.56 | 3.6468 | 1.199 | 5.15184 | 5.06235 | 1.3833 | 1.64 | 0.710016 | **0.69** |
| $b(m)$ | 0.127 | 0.12 | 0.1112 | 0.0924 | 0.1066 | 0.24725 | 0.24147 | 0.125 | 0.1054 | 0.154626 | **0.1526** |
| $d_o(m)$ | 0.025 | 0.02 | 0.015 | 0.0105 | 0.015 | 0.01204 | 0.01171 | 0.0100 | 0.01575 | 0.011441 | **0.0114** |
| $P_t(m)$ | 0.031 | 0.025 | 0.0187 | - | 0.0188 | 0.01505 | 0.01464 | 0.0125 | 0.01968 | 0.014301 | **0.0143** |
| $C_1(m)$ | 0.006 | 0.005 | 0.0037 | - | 0.0038 | 0.00301 | 0.00293 | 0.0025 | | 0.00286 | **0.0028** |
| $n_t$ | 4 | 4 | 2 | 2 | 2 | 1 | 1 | 2 | | 2 | **2** |
| $N_t$ | 158 | 391 | 646 | 511 | 1061 | 301 | 309 | 1152.888 | 924 | 635.2294 | **635.2587** |
| $v_t(m/s)$ | 1.44 | 0.87 | 0.93 | 0.43 | 0.69 | 0.8615 | 0.8871 | 0.6522 | 0.677 | 0.904129 | **0.9041** |
| $Re_t$ | 8227 | 4068 | 3283 | - | 2298 | 2306.77 | 2303.46 | 1450.0174 | 2408 | 2299.998 | **2299.9910** |
| $Pr_t$ | 55.2 | 55.2 | 55.2 | - | 55.2 | 56.4538 | 56.4538 | 56.4538 | 55.2 | 56.45385 | **56.4538** |
| $h_t(W/m^2K)$ | 619 | 1168 | 1205 | 2186 | 1251 | 1398.85 | 1435.68 | 1639.2213 | 1262 | 1174.574 | **1208.7230** |
| $f_t$ | 0.033 | 1168 | 0.044 | - | 0.05 | 0.04848 | 0.04854 | 0.0591 | 0.049 | 0.049861 | **0.04986** |
| $\Delta P_t(Pa)$ | 49245 | 14009 | 16926 | 1696 | 5109 | 10502.45 | 11165.45 | 5382.9311 | 9374 | 5179.414 | **5091.273** |
| $a_s(m^2)$ | 0.0137 | 0.0148 | 0.0131 | - | 0.0158 | 0.01585 | 0.01527 | 0.0114 | | 0.01237 | **0.0122** |
| $D_e(m)$ | 0.025 | 0.019 | 0.0149 | - | 0.0149 | 0.01188 | 0.01157 | 0.0071 | 0.0156 | 0.008134 | **0.0081** |
| $v_s(m/s)$ | 0.47 | 0.43 | 0.495 | 0.37 | 0.432 | 0.40948 | 0.42526 | 0.5672 | 0.4 | 0.524974 | **0.5316** |
| $Re_s$ | 25281 | 18327 | 15844 | - | 13689 | 10345.29 | 10456.39 | 8568.0357 | 14448 | 9073.644 | **9188.785** |
| $Pr_s$ | 7.5 | 7.5 | 7.5 | - | 7.5 | 7.6 | 7.6 | 7.6 | 7.5 | 7.6 | **7.6** |
| $h_s(W/m^2K)$ | 920 | 1034 | 1288 | 868 | 1278 | 1248.86 | 1290.789 | 2062.1966 | 1156 | 1857.576 | **1870.585** |
| $f_s$ | 0.315 | 0.331 | 0.337 | - | 0.345 | 0.35987 | 0.35929 | 0.3702 | 0.3422 | 0.367025 | **0.3663** |
| $\Delta P_s(Pa)$ | 24909 | 15717 | 21745 | 10667 | 15275 | 14414.26 | 15820.74 | 36090.0964 | 12768 | 9708.001 | **9780.794** |
| $U(W/m^2K)$ | 317 | 376 | 409.3 | 323 | 317.75 | 326.071 | 331.358 | 381.6827 | 347.6 | 336.1286 | **339.9925** |
| $S(m^2)$ | 61.5 | 52.9 | 47.5 | 61.566 | 60.35 | 58.641 | 57.705 | 50.09702 | 56.6 | 56.84084 | **56.1948** |
| $C_{inv}(€)$ | 19007 | 17599 | 16707 | 19014 | 18799 | 18536.55 | 18383.46 | 17129.8543 | 18202 | 18241.79 | **18135.82** |
| $C_{annual}(€/year)$ | 1304 | 440 | 523.3 | 197.139 | 164.414 | 272.576 | 292.7937 | 352.885 | 210.2 | 155.71 | **154.3616** |
| $C_{total\_disc}(€)$ | 8012 | 2704 | 3215.6 | 1211.3 | 1010.25 | 1674.86 | 1799.09 | 2163.3257 | 1231 | 956.79 | **948.485** |
| $C_{total}(€)$ | 27020 | 20303 | 19922.6 | 20225 | 19810 | 20211 | 20182 | 19298.18 | 19433 | 19198.58 | **19084.31** |

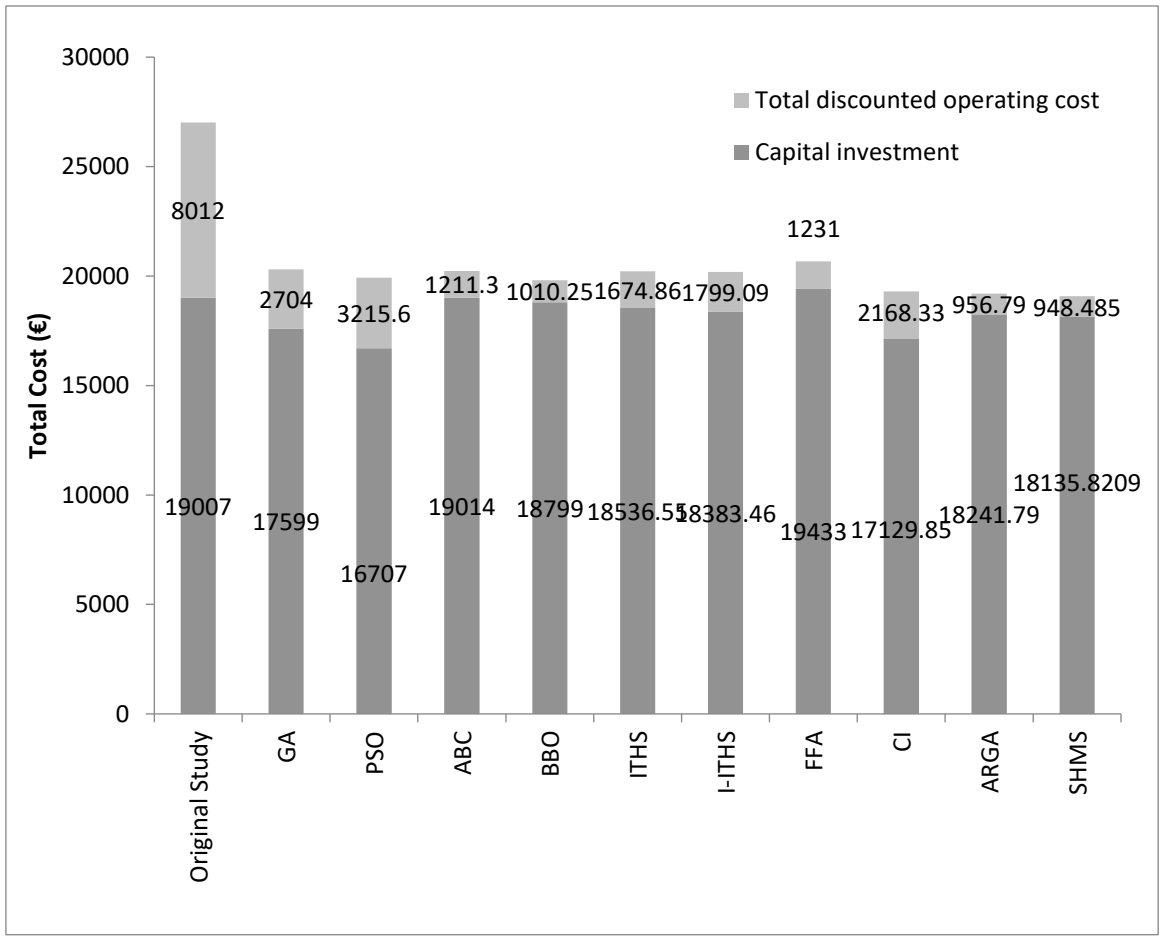

Fig. 8: Total Cost Comparison for Case 2

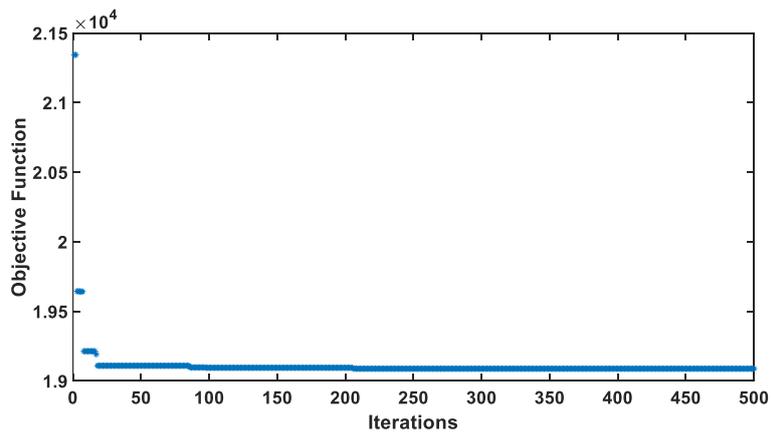

Fig. 9: Convergence plot for Design and Economic Optimization of Shell and Tube Heat Exchanger Problem Case 2

**Table 22: Comparison of SHMS results Design and Economic Optimization of Shell and Tube Heat Exchanger Problem Case 3**

| Parameters | Original Study | GA | PSO | ABC | BBO | ITHS | I-ITHS | CI | TLBO | SAMPE-Jaya | ARGA | SHMS |
|---|---|---|---|---|---|---|---|---|---|---|---|---|
| $D_s(m)$ | 0.387 | 0.62 | 0.0181 | 1.0024 | 0.55798 | 0.5726 | 0.5671 | 0.5235 | 0.5524 | 0.5671 | 0.460204468 | **0.4702** |
| $L(m)$ | 4.88 | 1.548 | 1.45 | 2.4 | 1.133 | 0.9737 | 0.9761 | 1.1943 | 0.9854 | 0.9569 | 0.793852708 | **0.7054** |
| $b(m)$ | 0.305 | 0.44 | 0.423 | 0.354 | 0.5 | 0.4974 | 0.4989 | 0.5000 | 0.464 | 0.499 | 0.460204468 | **0.5104** |
| $d_o(m)$ | 0.019 | 0.016 | 0.0145 | 0.103 | 0.01 | 0.0101 | 0.01 | 0.0100 | 0.010 | 0.01 | 0.011981248 | **0.01** |
| $P_t(m)$ | 0.023 | 0.02 | 0.0187 | - | 0.0125 | 0.0126 | 0.0125 | 0.0125 | 0.0125 | 0.0125 | 0.01497656 | **0.0125** |
| $C_1(m)$ | 0.004 | 0.004 | 0.0036 | - | 0.0025 | 0.0025 | 0.0025 | 0.0025 | | | 0.002995312 | **0.0025** |
| $n_t$ | 2 | 2 | 2 | 2 | 2 | 2 | 2 | 2 | | | 2 | **2** |
| $N_t$ | 160 | 803 | 894 | 704 | 1565 | 1845 | 1846 | 1548.6665 | 1743 | 1841 | 781.7678209 | **1222.003** |
| $v_t(m/s)$ | 1.76 | 0.68 | 0.74 | 0.36 | 0.898 | 0.747 | 0.761 | 0.9083 | 0.80695 | 0.76399 | 1.25317137 | **1.1508** |
| $Re_t$ | 36409 | 9487 | 9424 | - | 7804 | 6552 | 6614 | 7889.7151 | 7009.98 | 6636.82 | 13043.08036 | **9997.4150** |
| $Pr_t$ | 6.2 | 6.2 | 6.2 | - | 6.2 | 6.2 | 6.2 | 6.2026 | 6.2026 | 6.2025 | 6.202580645 | **6.2025** |
| $h_t(W/m^2K)$ | 6558 | 6043 | 5618 | 4438 | 9180 | 5441 | 5536 | 4901.7267 | | | 6290.111612 | **6170.5740** |
| $f_t$ | 0.023 | 0.031 | 0.0314 | - | 0.0337 | 0.0369 | 0.0368 | 0.0336 | 0.034817 | 0.035386 | 0.029220623 | **0.0314** |
| $\Delta P_t(Pa)$ | 62812 | 3673 | 4474 | 2046 | 4176 | 3869 | 4049 | 6200.0472 | 4416.42 | 3926.01 | 7719.023019 | **6975.8420** |
| $a_s(m^2)$ | 0.0236 | 0.0541 | 0.059 | - | 0.0558 | 0.0569 | 0.0565 | 0.0523 | 5789.17 | 5541.30 | 0.04235763 | **0.0480** |
| $D_e(m)$ | 0.013 | 0.015 | 0.01 | - | 0.0071 | 0.0071 | 0.0071 | 0.0071 | 0.00711 | 0.00711 | 0.008517657 | **0.0071** |
| $v_s(m/s)$ | 0.94 | 0.41 | 0.375 | 0.12 | 0.398 | 0.3893 | 0.3919 | 0.4237 | 0.4326 | 0.39172 | 0.523657822 | **0.4620** |
| $Re_s$ | 16200 | 8039 | 4814 | - | 3515 | 3473 | 3461 | 3746.0280 | 3830.527 | 3467.839 | 5547.544747 | **4085.1680** |
| $Pr_s$ | 5.4 | 5.4 | 5.4 | - | 5.4 | 5.4 | 5.4 | 5.3935 | 5.3935 | 5.3935 | 5.393548387 | **5.3935** |
| $h_s(W/m^2K)$ | 5735 | 3476 | 4088.3 | 5608 | 4911 | 4832 | 4871 | 5078.1022 | 5374.56 | 5088.428 | 5267.295773 | **5333.3460** |
| $f_s$ | 0.337 | 0.374 | 0.403 | - | 0.423 | 0.4238 | 0.4241 | 0.4191 | 0.4177 | 0.423988 | 0.395136701 | **0.4136** |
| $\Delta P_s(Pa)$ | 67684 | 4365 | 4271 | 27166 | 5917 | 4995 | 5062 | 6585.2425 | 6412.95 | 4928.072 | 5024.067328 | **4016.2380** |
| $U(W/m^2K)$ | 1471 | 1121 | 1177 | 1187 | 1384 | 1220 | 1229 | 1198.4141 | 1274.73 | 1242.84 | 1296.89011 | **1294.3750** |
| $S(m^2)$ | 46.6 | 62.5 | 59.2 | 54.72 | 55.73 | 57.3 | 56.64 | 58.0975 | 53.9355 | 55.318 | 59.48776644 | **59.6033** |
| $C_{inv}(€)$ | 16549 | 19163 | 18614 | 17893 | 18059 | 18273 | 18209 | 18447.6373 | 17764.30 | 17991.96 | 18674.91 | **18693.7960** |
| $C_{annual}(€/year)$ | 4466 | 272 | 276 | 257.82 | 203.68 | 231 | 238 | 383.4699 | 278.455 | 231.53 | 346.19 | **333.7221** |
| $C_{total\_disc}(€)$ | 27440 | 1671 | 1696 | 1584.2 | 1251.5 | 1419 | 1464 | 2356.2566 | 1710.988 | 1422.69 | 2127.18 | **2050.5779** |
| $C_{total}(€)$ | 43989 | 20834 | 20310 | 19478 | 19310 | 19693 | 19674 | 20803.8940 | 19475.297 | 19414.65 | 20802.09 | **20744.3639** |

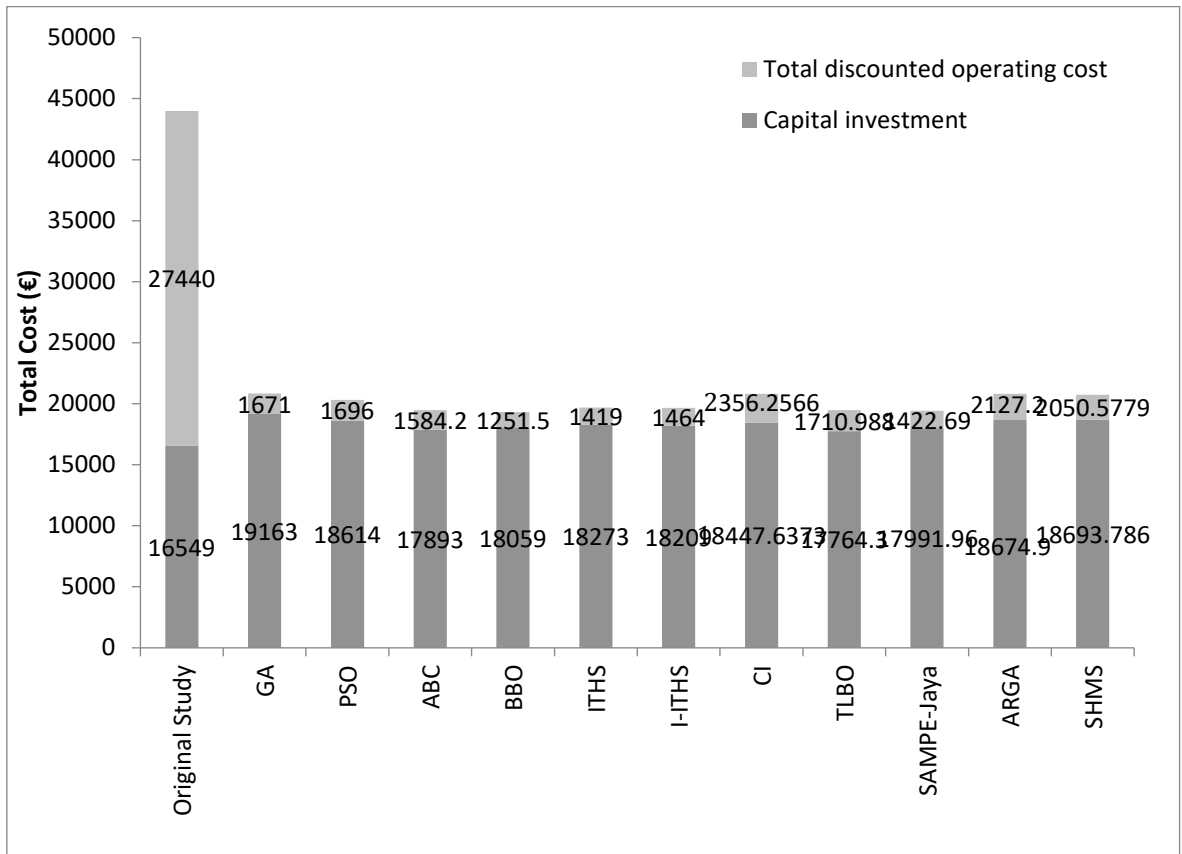

**Fig. 10: Total Cost Comparison for Case 3**

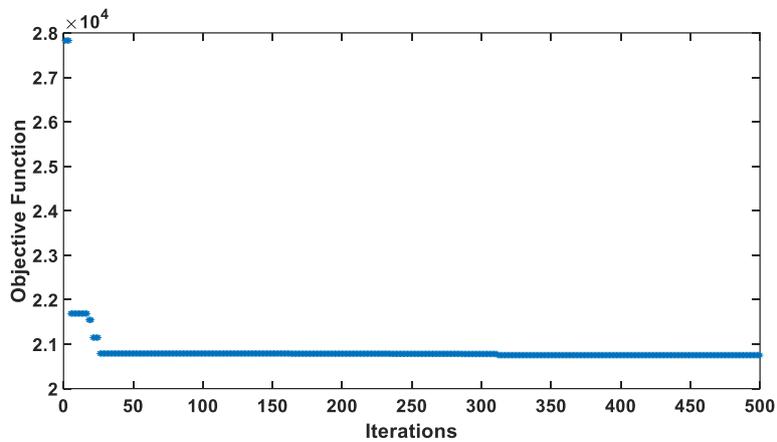

**Fig. 11: Convergence plot for Design and Economic Optimization of Shell and Tube Heat Exchanger Problem Case 3**

Table 23: SHMS performance details

| Case studies | Solutions | | Standard Deviation | Avg. No. of Function Evaluations (FE) | Avg. Comp. Time(sec) |
|---|---|---|---|---|---|
| | Best | | | | |
| | Mean | | | | |
| | Worst | | | | |
| Case 1 | 41718.6558 | | 4.0847 | 20510 | 9.62 |
| | 41725.3892 | | | | |
| | 41728.6558 | | | | |
| Case 2 | 19084.3059 | | 3.1663 | 17235 | 8.10 |
| | 19088.3476 | | | | |
| | 19097.2054 | | | | |
| Case 3 | 20744.3639 | | 1.4565 | 44721 | 20.13 |
| | 20746.1280 | | | | |
| | 20749.8314 | | | | |

Table 24: Closeness of SHMS solutions with other algorithms

| Case studies | Referred algorithms | Solutions of Total Cost (€) | Closeness to the Best Reported Solution (%) |
|---|---|---|---|
| Case 1 | Original Study | 64480 | **35.2998 ↑** |
| | GA | 55077 | **24.2539 ↑** |
| | PSO | 53231.1 | **21.6272 ↑** |
| | ABC | 50793 | **17.8653 ↑** |
| | BBO | 50582 | **17.5227 ↑** |
| | ITHS | 50226 | **16.9381 ↑** |
| | I-ITHS | 50173 | **16.8503 ↑** |
| | FFA | 45783 | **8.8774 ↑** |
| | CI | 50006.18 | **16.5729 ↑** |
| | TLBO | 45,782 | **8.8754 ↑** |
| | SAMPE-Jaya | 42015.98 | **0.7076 ↑** |
| | ARGA | 41913.54 | **0.4649 ↑** |
| Case 2 | Original Study | 27020 | **29.3697 ↑** |
| | GA | 20303 | **6.0025 ↑** |
| | PSO | 19922.6 | **4.2077 ↑** |
| | ABC | 20225 | **5.6400 ↑** |
| | BBO | 19810 | **3.6632 ↑** |
| | ITHS | 20211 | **5.5746 ↑** |
| | I-ITHS | 20182 | **5.4389 ↑** |
| | FFA | 19433 | **1.7943 ↑** |
| | CI | 19298.18 | **1.1082 ↑** |
| | ARGA | 19198.58 | **0.5952 ↑** |
| Case 3 | Original Study | 43989 | **52.8419 ↑** |
| | GA | 20834 | **0.4302 ↑** |
| | PSO | 20310 | 2.1386 ↓ |
| | ABC | 19478 | 6.5015 ↓ |
| | BBO | 19310 | 7.4280 ↓ |
| | ITHS | 19693 | 5.3387 ↓ |
| | I-ITHS | 19674 | 5.4405 ↓ |

| | CI | 20803.89 | **0.2861 ↑** |
| | TLBO | 19475.297 | 6.5162 ↓ |
| | SAMPE-Jaya | 19414.65 | 6.8490 ↓ |
| | ARGA | 20802.09 | **0.2775 ↑** |

## Conclusion and Future Directions

In this paper, a novel Snail Homing and Mating Search (SHMS) algorithm inspired from the living habitat of the snails is proposed. Snails usually live in moist and humid regions and continuously travels to find food and a mate, leaving behind a trail of mucus that serves as a guide for their return journey. Snails tend to navigate by following the available trails on the ground and responding to cues from nearby shelter homes. The approach is validated by solving set of benchmark problems consisting of 2 different groups of unimodal (UM)and multi-modal (MM) functions(F1-F23). To check the SHMS algorithms scalability, F1-F13 problems are solved for 30, 100, 500 and 1000 dimensions. Wilcoxon and Friedman statistical tests are conducted for comparing the performance of the SHMS algorithm with the existing algorithms. The performance of SHMS algorithm is exceedingly better as compared to PSO, GWO, FFA, WOA, TLBO, MFO, BBO, DE, SSA, GSA and IPO while solving F1-F13 functions for 30, 100, 500 and 1000 dimensions in terms of objective function value (best and mean), robustness, as well as computational time. The performance of the SHMS algorithm is marginally better as compared to AVOA for 30 and 100 dimensions. The solution quality highlighted that the SHMS algorithm is a robust approach with reasonable computational cost and could quickly reach in the close neighbourhood of the global optimum solution. For fixed dimensions F14-F23 functions SHMS algorithm could not yield better solutions as compared to other algorithms due to weak exploitation. In addition, real-world application of SHMS algorithm is successfully demonstrated in the engineering design domain by solving three cases of shell and tube heat exchanger problem. The total cost ($C_{total}$) is significantly minimized. The results are compared with other well-known metaheuristic algorithms such as GA, PSO, ABC, BBO, ITHS, I-ITHS, FFA, CI, TLBO, SAMPE-Jaya and ARGA. A generalised constraint handling mechanism needs to be developed and incorporated into the SHMS algorithm. This can help SHMS algorithm to solve real-world problems which are generally constrained in nature. The constrained SHMS version could be further extended to solve complex structural optimisation problems.